\documentclass[lettersize,journal]{IEEEtran}
\usepackage{amssymb}
\usepackage{amsmath}
\usepackage{amsfonts}
\usepackage{algorithmic}
\usepackage{algorithm}
\usepackage{array}
\usepackage[caption=false,font=normalsize,labelfont=sf,textfont=sf]{subfig}
\usepackage{textcomp}
\usepackage{stfloats}
\usepackage{url}
\usepackage{verbatim}
\usepackage{graphicx}
\usepackage{cite}
\usepackage{float}
\usepackage{multirow} 
\usepackage{booktabs} 
\usepackage{array}
\usepackage{color}
\usepackage{xcolor}
\usepackage{caption}
\usepackage{epstopdf} 
\usepackage{subfiles} 
\usepackage{hyperref} 
\definecolor{darkred}{RGB}{170, 0, 0}
\newcommand{\modified}[1]{#1}
\begin{document}

\title{Multi-Source Urban Traffic Flow Forecasting\\with Drone and Loop Detector Data}

\author{
Weijiang Xiong,
Robert Fonod~\IEEEmembership{Senior Member~IEEE},\\
Alexandre Alahi~\IEEEmembership{Member~IEEE},
Nikolas Geroliminis~\IEEEmembership{Member~IEEE}
\thanks{Manuscript received MM DD, YYYY; revised MM DD, YYYY. 
This work has been supported by the Swiss National Science Foundation (SNSF) under NCCR Automation, grant agreement number 51NF40\_180545.
\textit{(Corresponding author: Nikolas Geroliminis.)}
}
\thanks{Weijiang Xiong, Robert Fonod and Nikolas Geroliminis are with the Urban Transport Systems Laboratory (LUTS), École Polytechnique Fédérale de Lausanne (EPFL), Switzerland. (email: weijiang.xiong@epfl.ch; robert.fonod@epfl.ch; nikolas.geroliminis@epfl.ch)}
\thanks{Alexandre Alahi is with the Visual Intelligence for Transportation Laboratory (VITA), École Polytechnique Fédérale de Lausanne (EPFL), Switzerland. (email: alexandre.alahi@epfl.ch)}
}

\markboth{
Submission to IEEE Transactions on Intelligent Transportation Systems
}%
{
}


\maketitle

\begin{abstract}
Traffic forecasting is a fundamental task in transportation research, however the scope of current research has mainly focused on a single data modality of loop detectors.
Recently, the advances in Artificial Intelligence and drone technologies have made possible novel solutions to efficient, accurate and flexible aerial observations of urban traffic.
As a promising traffic monitoring approach, drones can create an accurate multi-sensor mobility observatory for large-scale urban networks, when combined with existing infrastructure.
Therefore, this paper investigates urban traffic prediction from a novel perspective, where multiple input sources, i.e., loop detectors and drones, are utilized to predict both the individual traffic of all road segments and the regional traffic.
A simple yet effective graph-based model, \modified{Hierarchical Multi-Source Neural Network (HiMSNet)}, is proposed to integrate multiple data modalities and learn spatio-temporal correlations. 
Detailed analysis shows that predicting accurate segment-level speed is more challenging than the regional speed, especially under high-demand scenarios with heavier congestions and varying traffic dynamics.
Utilizing both drone and loop detector data, the prediction accuracy can be improved compared to single-modality cases, when the sensors have lower coverages and are subject to noise.
Our simulation study based on vehicle trajectories in a real urban road network has highlighted the value of integrating drones in traffic forecasting and monitoring.
\end{abstract}


\section{Introduction}
\label{sec: introduction}

Knowing the current traffic states and their future evolution is essential for traffic management and control in Intelligent Transportation Systems (ITS).
Therefore, in the literature, traffic forecasting has been a primal topic of many research works.
In practice, traffic monitoring and forecasting has been an important tool for transportation policy and traffic management.
With accurate predictions, road users can dynamically adjust their route choices to avoid congested areas~\cite{liebig2017dynamic, zeng2021IVS}.
City-wide traffic flow predictions can support traffic signal control~\cite{jiang2021urban}, plannings of platoons~\cite{chaanine2024real}, dynamic transportation system management~\cite{amini2016traffic} and the optimal control of autonomous buses~\cite{pasquale2022traffic}.
Time-crucial services, such as emergency response, can also schedule their operations based on the predicted traffic to guarantee the arrival time~\cite{nguyen2021controllable}.
These applications will improve the overall efficiency of the transportation network, reduce energy consumption, and create greater value for the economy and sustainability of society.

Owing to the limitations of data availability, many deep learning methods are built upon highway loop detector data, e.g., METR-LA and PEMS-Bay~\cite{li2018dcrnn}. 
However, urban traffic forecasting presents additional challenges due to the dense and complex road network structure.
Recently, Unmanned Aerial Vehicles (UAVs) or drones are emerging as a novel information-gathering solution in ITS~\cite{butilua2022urban}, as they can flexibly fly over an area of interest and record high-quality videos. 
At the same time, the growing power of Computer Vision, Artificial Intelligence and computational resources have made it possible to automatically and efficiently extract vehicle locations from the captured videos. 
These advantages have empowered drones to collect rich and accurate information from a large spatial region, making them a preferable choice for urban traffic monitoring. 

\modified{Specifically, Barmpounakis et~al.~\cite{barmpounakis2020new} have conducted a large-scale experiment in Athens, where a swarm of drones was deployed to monitor traffic in the city center.
The drone-captured videos can support in-depth analysis and understanding of urban traffic, because they present clear views of the monitored areas, and most vehicles can be identified by object detection and tracking methods~\cite{fonod2024advanced}.}
With the detailed vehicle trajectories, the understanding of various traffic phenomena can be significantly improved, including emissions~\cite{espadaler2023traffic}, multi-modal \modified{Macroscopic Fundamental Diagrams (MFD)}~\cite{paipuri2021empirical}, path flow estimation~\cite{cao2024tracking}, car following~\cite{hart2024towards}, traffic signal safety~\cite{ashqer2024evaluating} and parking monitoring~\cite{kim2024monitoring}.
With vehicle trajectories, traffic variables can be accurately computed for all visible roads in the field of view (FOV)~\cite{barmpounakis2019accurate}, resulting in a great advantage compared to grounded sensors, e.g., loop detectors and \modified{video suveillance cameras}, whose scopes are limited to the roads with installed facilities.

The high-quality data from drones can provide solid grounds for traffic state prediction, which will be an additional asset to the existing traffic forecasting methods based on loop detector data. 
However, few studies have attempted to address the traffic forecasting problem using joint observatories from drones and other information sources.
\modified{Besides, little attention has been paid to deep learning-based regional traffic predictions, which is an essential task for urban traffic management.}
Therefore, this paper aims to investigate a promising paradigm of multi-source urban traffic forecasting, where drone-measured traffic are combined with loop detector data to provide high-quality predictions \modified{for both individual road segments and urban regions at the same time}.

In summary, the main contributions are as follows.
\modified{First, this paper introduces a flexible deep learning model HiMSNet to jointly handle road segment-level and regional traffic prediction, using traffic measurements of simulated drones and loop detectors.}
Second, A well-calibrated multi-source dataset SimBarca is created, offering simulated traffic speed data of drones and loop detectors in Barcelona. 
\modified{Using the simulated trajectories of all vehicles in a dense urban road network, SimBarca introduces the prediction tasks of segment-level and regional traffic speeds.}
Third, \modified{extensive} experiments and analysis are provided to evaluate HiMSNet on SimBarca, \modified{highlighting the necessity of drone data in accurate urban traffic prediction.}
\modified{Finally, the data, model and evaluations} in this work are openly shared with the community to facilitate future research. (Link will be provided upon acceptance)

\section{Related Work}
\label{sec: related_work}

The multi-source traffic forecasting problem is an extension of the traditional traffic forecasting task. 
As a popular topic in transportation research, traffic forecasting has a long history and a vast collection of literature.
Over the past years, the development of this field has experienced a paradigm shift from classical statistical approaches to data-driven approaches~\cite{vlahogianni2014short, lv2014traffic}.
Despite the great methodological differences, the critical challenge of the traffic forecasting problem remains, which is the complex temporal dynamics and spatial correlations in the data.
Therefore, most deep learning methods will design specific neural network modules to tackle the dependencies in spatial and temporal dimensions.

\subsection{GNN-Based Methods}
\label{subsec:gnn based methods}

Since the road network has a natural graph structure, Graph Neural Networks (GNN) have been extensively explored in traffic forecasting~\cite{jin2023spatio}.
\modified{In a GNN-based model, nodes usually represent sensors (e.g., loop detectors) at various locations, and the edges correspond the road connections between various locations.
The model will gather information from the graph neighbors of each node, which are often the other sensors within a dinstance.
As a result, GNNs can pass messages according to the graph structure and thus learn spatial relations.}

In the modern era of deep learning, the Diffusion Convolutional Recurrent Neural Network (DCRNN)~\cite{li2018dcrnn} is one of the pioneering works to utilize a graph-based model in this area.
In DCRNN, the temporal dependencies are learned by a Recurrent Neural Network (RNN), whose layer has an embedded diffusion convolution module to model the spatial dependencies by a diffusion process in the sensor graph.
A network block in Spatio-temporal Graph Convolutional Neural Networks (STGCN)~\cite{yu2018STGCN} applies a convolutional operation over the time dimension, and then propagates the information over the spatial graph structure with Graph Convolution.
Following a similar design philosophy, the Temporal Graph Convolutional Network (T-GCN)~\cite{zhao2019tgcn} propagates neighborhood information over the graph at each time step and then applies an RNN to learn the temporal dependencies.

However, the correlations in the data are often beyond the description of a pre-defined graph structure, which is typically an adjacency matrix based on geographical distance or road topology.
Therefore, many subsequent works have focused on learning an adaptive graph structure from the data, instead of following a fixed graph.
Graph WaveNet~\cite{wu2019graph} develops an adaptive graph to learn the spatial relations between the sensors, which is achieved by composing the graph with learnable positional embeddings.
The Attention-based Spatial-Temporal Graph Convolutional Networks (ASTGCN)~\cite{guo2019attention} further introduces an attention-based mechanism to both the spatial and temporal dimensions, which effectively assigns different weights to the neighbors according to the correlation patterns.
The Adaptive Graph Convolutional Recurrent Network (AGCRN)~\cite{bai2020adaptive} also employs an embedding-based adaptive graph as Graph WaveNet~\cite{wu2019graph}, but proposes to learn a normalized adjacency matrix instead of the adjacency itself.

This spatio-temporal framework has been further developed in the literature.
In Multi-variate Time Series Graph Neural Network (MTGNN)~\cite{wu2020connecting}, a node will exchange information with neighborhoods at different distances, using the Mixhop~\cite{abu2019mixhop} technique. 
The Graph Multi-Attention Network (GMAN)~\cite{zheng2020gman} has an encoder-decoder structure equipped with an attention mechanism for space and time dimensions, and the encoder message is further enhanced by cross attention between historical and future time step embeddings. 
TwoResNet~\cite{Li2022tworesnet} addresses regional traffic with a low-resolution module at macroscopic scale, and the microscopic traffic is handled by a high-resolution module. 
The Traffic Graph Convolution (TGC) network~\cite{cui2019traffic} adopted intuition from transportation and defined the adjacency matrix based on the travel time between the spatial locations, instead of geographical distance.
The Multi-Weight Traffic Graph Convolution (MW-TGC) network~\cite{shin2020incorporating} integrated various metadata into the design of adjacency matrix, including geographical distance, angle, speed limit and number of paths between locations.

\subsection{Transformer-Based Methods}
\label{subsec:transformer based methods}

In recent years, the Transformer~\cite{vaswani2017attention} model has also been introduced into traffic forecasting.
The core mechanism of Transformer is attention, which computes pair-wise similarities between two sets of elements and then aggregates the information based on the similarities.
Such a generic design makes Transformers suitable for the adaptive learning of both spatial and temporal dependencies in traffic forecasting problem. 
Thus, this has become a shared foundation for many transformer-based methods. 

Yan et al.~\cite{yan2022TITS} use a transformer model to learn the spatial dependencies, and employ an \modified{Long Short-Term Memory Network (LSTM)}~\cite{hochreiter1997long} for temporal learning.
The adjacency matrix is used in transformer decoder part to enforce the spatial constraints.
Xu et al.~\cite{xu2020spatial} propose to learn temporal relations with transformers, and fuse the output of transformer and Graph Neural Network (GNN) to obtain a joint representation of the spatial relations.
In this way, the model can make use of the flexibility of pair-wise correlations in the transformer model, and also follow the spatial dependencies of the graph structure.
Similar to GMAN~\cite{zheng2020gman}, the Meta Graph Transformer (MGT)~\cite{ye2022mgt} designs a cross-attention between encoder and decoder, which allows the model to capture the relations between historical and future time steps.
To enhance the graph structure, MGT also encoded multiple weight matrices based on geographical distance, historical traffic flow and vehicle counts.
Shao et al.~\cite{shao2022step} adapted the method from Masked Auto-Encoder~\cite{he2022masked}, and propose to obtain a better data representation by a pretraining phase on multiple datasets, where a transformer model is trained to complete the missing values with only partial inputs.
PDFormer~\cite{jiang2023pdformer} clusters segments of traffic time series into sets of representative patterns, and utilize them to identify similar patterns in new input sequences when making predictions.

\subsection{Research Gap and Contribution}
\label{subsec:research gap and contribution}
\modified{
Although traffic forecasting has been widely studied, the scope of existing research has been centered on predicting loop detector-measured highway traffic, due to the limitation of data availability.
Aside from single-sensor traffic prediction, some recent research works have explored using multiple data sources to estimate urban traffic states using data combination~\cite{tak2014real, jiang2017traffic, essien2019improving}, probabilistic filtering~\cite{
nantes2016real, zhu2018urban}, model ensemble~\cite{koesdwiady2016improving} and deep feature fusion~\cite{zhao2018travel}.
However, the scope of these research works is limited to one or a few roads.}

In general, traffic forecasting datasets do not have big scales, and thus building complicated models is not always necessary.
For example, STID~\cite{shao2022spatial} achieves similar accuracy as the transformer-based PDFormer~\cite{jiang2023pdformer} by using Multi-Layer Perceptrons (MLPs) only.
A later study on time series forecasting has also shown that linear models can achieve similar performance as transformer models~\cite{zeng2023transformers} if properly tuned.
Therefore, this work will refrain from using a complex model design and will keep the implementation concise, as the primary aim is to provide a simple yet effective model along with comprehensive evaluations to build a solid foundation for future research.

\modified{
Compared to existing literature, this study expands the research scope in problem scale, data quality and prediction tasks.
First, this work aims to predict the traffic of all roads in an urban area with dense connections and highly dynamic traffic.
Secondly, the drone-measured traffic can accurately represent the traffic states of complete roads, instead of a short range around the installed loop detectors.
Finally, regional traffic prediction, has been made possible provided that the drones can coverage a reasonable percentage of the area.}

\section{Methodology}
\label{sec: methodology}

This section will first introduce the formulation of multi-source traffic forecasting (notation inspired by~\cite{horn2020set}).
Then a baseline model for multi-source traffic forecasting will be presented, and the final part will describe the dataset creation method using vehicle trajectories.

\subsection{Multi-Source Traffic Forecasting}

In a modern transportation network, there are various types of sensors (information sources) providing traffic information at different locations in different ways.
If the network traffic states $\mathbf{X}$ are observed at time $t$, location $p$ using method $m$, and the measured value is $v$, this measurement procedure can be described as 
\begin{equation}
v = \mathcal{O}_m (\mathbf{X}, t, p),
\end{equation}
where $\mathcal{O}_m$ represents the general physical or statistical process to obtain the measurement. 

For a real-world urban transportation network, the accurate and true state $\mathbf{X}$ is often unknown or difficult to determine.
The practically available information is what the sensors provide, for example, one such measurement can be noted as $(t, v, m, p)$.
Therefore, the observations collected over time at location $p$ using method $m$ can be noted as a time series:
\begin{equation}
\mathcal{S}^{m,p} = \left\{ (t_0, v_0), (t_1, v_1), ..., (t_n, v_n)~|~m, p\right\},
\end{equation}
where the time steps $(t_0, t_1, ..., t_n)$ are chronologically sorted.
Such a time series has its own evolution patterns over time, and at the same time is correlated with time series from other locations, since they belong to the same transportation network $\mathcal{G}$.
Besides, since different observation methods reveal different aspects of the traffic states, i.e., different sensor coverages, the inter-modality correlation is also important in multi-source traffic forecasting.
As a result, if there are $M$ observation methods and $P$ locations in total, a Multi-Source Time Series (MSTS) is the collection of time series from all modalities and locations, which comprehensively describe the available traffic information in the transportation network.
\begin{equation}
\modified{\mathcal{S} = \left\{ \mathcal{S}^{m,p} \right\}, \text{where}~m\in [1, 2, ..., M],~p\in [1, 2, ..., P] .}
\end{equation}

An MSTS is considered synchronized if, for any time step with records, the measurements with all modalities on the same variable are available at all locations.
This will be the special case where all the road segments of interest are covered by all types of sensors, and these sensors are designed to provide measurements at the same time steps.
However, this is not realistic mainly because different sensors can have different update strategies and spatial coverages.
As a result, the MSTS for a transportation network is usually unsynchronized, and it cannot be simply concatenated to be a time series with multiple variables.
Instead, the spatial correlations and modality characteristics should be considered in the model.

With this concept, a dataset $\mathcal{D}$ is a collection of MSTS $\mathcal{S}_k$, the corresponding labels $\mathbf{y}_k$ and the road graph $\mathcal{G}$, with $k$ being the index of the sample.
\begin{equation}
\mathcal{D} = \{(\mathcal{S}_1, \mathbf{y}_1), (\mathcal{S}_2, \mathbf{y}_2), ..., (\mathcal{S}_N, \mathbf{y}_N); \mathcal{G}\}.
\end{equation}
\modified{Following the same notation convention, $\mathcal{S}_k^{m, p}$ will be the traffic variable measured by method $m$ at location $p$ in the $k$-th sample.
And the details for creating the dataset and the samples will be presented in Section~\ref{subsec: dataset creation}.}

Suppose a neural network $f$ parameterized by $\theta$ is trained to predict $\mathbf{y}$ using the input MTMS $\mathcal{S}$ and the road graph $\mathcal{G}$, the training loss can be generally noted as: 
\begin{equation}
L(\theta, \mathcal{D}) = 
\mathbb{E}_{(\mathcal{S}, \mathbf{y}) \in \mathcal{D}}\left[l(\mathbf{y}; f_\theta(\mathcal{S}, \mathcal{G})) \right],
\end{equation}
where $l$ is a distance function to measure the differences between predictions and labels.
\modified{Specifically, both regional and segment-level speed prediction use the Mean Absolute Error (MAE, as detailed in Equation~\ref{equ: mean prediction evaluation metrics}) as the loss function.
In case no vehicle is observed, the traffic variable is marked as missing, and will be ignored in loss calculation and performance evaluation.
The overall training loss $l$ for a batch is a sum of the segment-level loss $l_{seg}$ and the regional loss $l_{reg}$:
\begin{equation}
    l = l_{seg} + l_{reg} 
\end{equation}}

\subsection{Model Architecture}
\label{subsec: model architecture}

\subsubsection{Overall Pipeline}
\begin{figure*}[ht]
    \centering
    \includegraphics[width=0.8\textwidth]{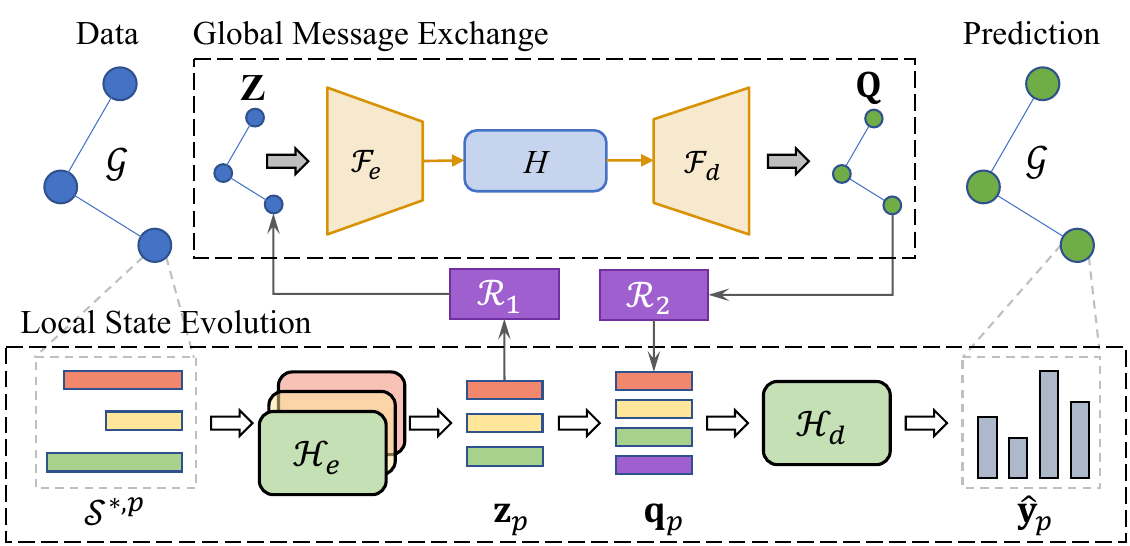}
    \caption{The two-layer model structure of HiMSNet. 
    \modified{The local layer processes all available time series data of a road segment, and concatenates the feature vectors of different data modalities into a joint feature}.
    The global layer exchanges messages across the road network, and enhance the local joint features with vicinity information.}
    \label{fig: model architecture}
\end{figure*}

Figure~\ref{fig: model architecture} shows the structure of our baseline model Hierarchical Multi-Source Network (HiMSNet) for traffic forecasting.
This architecture consists of a Global Message Exchange (GME) module to learn the spatial correlations between different locations, and a Local State Evolution (LSE) module to handle temporal dependencies and multiple data modalities at each location.
It is a simple and straightforward model that addresses the spatial, temporal and modality correlations in multi-source traffic forecasting.

Following the road network, the input data can be naturally organized as a graph structure, where a node corresponds to a road segment, and an edge indicates the connection between two road segments.
\modified{Before feeding the data into the deep learning model, all data modalities are normalized to have zero mean and unit variance.}
In the LSE module, the temporal encoder $\mathcal{H}_e$ will take $\mathcal{S}^{*,p}$, i.e., the time series from all observation methods at location $p$ (modalities marked in colors), and separately encode the time series of each modality into a unified representation $\mathbf{z}_p$.~\footnote{\modified{An encoder is a neural network module that transforms the input data to a hidden state, and a decoder computes desired outputs from the hidden state.}}
The LSE module will then concatenate the representations of all modalities to form a joint feature $\mathbf{z}_p$ for location $p$.
Then, $\mathbf{z}_p$ will go through a message encoding layer $\mathcal{R}_1$ and become a message to be shared with other locations.
The GME module will then take the messages from all locations ($\mathbf{Z}$), use its encoder $\mathcal{F}_e$ and decoder $\mathcal{F}_d$ to compute the new feature with exchanged information ($\mathbf{Q}$).
The message decoding layer $\mathcal{R}_2$ will take the element in $\mathbf{Q}$ corresponding to location $p$, and concatenate it together with the elements in $\mathbf{z}_p$ to form a joint feature $\mathbf{q}_p$, which contains the information from the location $p$ itself and its vicinity. 
Finally, the temporal decoder $\mathcal{H}_d$ will take $\mathbf{q}_p$ and predict the future traffic states $\hat{\mathbf{y}}_p$.

\begin{figure}[ht]
    \centering
    \includegraphics[width=0.48\textwidth]{"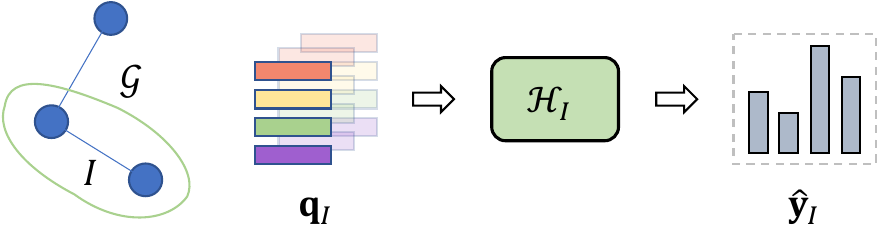"}
    \caption{Regional prediction branch. \modified{This branch groups the segment-level features by spatial regions (Figure~\ref{fig: region cluster and grid}), averages the features by group, and use them to predict regional traffic.}} 
    \label{fig: regional branch}
\end{figure}

HiMSNet is flexible and can be easily extended with an additional branch (Figure~\ref{fig: regional branch}) for regional traffic forecasting, which can be helpful in regional traffic control.
Given a defined region $I$, the regional branch will take the average of joint feature $\mathbf{q}_p$ for all $p \in I$ as the regional representation $\mathbf{q}_I$, and predict the regional average traffic state $\hat{\mathbf{y}}_I$ with the prediction module $\mathcal{H}_I$.
\modified{The components related to this branch are in parallel with the right part of LSE module in Figure~\ref{fig: model architecture}, i.e., $\mathbf{q}_p$, $\mathcal{H}_d$ and $\hat{\mathbf{y}}_p$, and regional features are averaged from the segment-level features belonging to the region.}

\subsubsection{Building Blocks and Implementation}

\modified{In HiMSNet, the building blocks for temporal and spatial modeling are LSTM~\cite{hochreiter1997long} and GCNConv~\cite{kipf2016semi} respectively.}

\modified{LSTM~\cite{hochreiter1997long} is a type of RNN with long-term modeling capabilities, and is widely used in sequential data analysis, such as time series.
At each time step, LSTM~\cite{hochreiter1997long} uses the current input $\mathbf{x}_t$, previous hidden state $\mathbf{h}_{t-1}$ and previous cell state $\mathbf{c}_{t-1}$ to compute the three internal gates $i$, $f$ and $g$, the output $\mathbf{o}_t$, new hidden state $\mathbf{h}_t$ and cell state $\mathbf{c}_t$.
The three internal gates and the output have individual learnable parameters for input and hidden states.
For example, in the $g$ gate $\mathbf{W}_{ig}$ is the weight matrix for the input, and $\mathbf{W}_{hg}$ is for hidden state (the notations for biases $\mathbf{b}$ are similar).
The computation can be expressed using the set of formulas in Equation~\ref{equ: lstm}, where $\sigma$ is the sigmoid function, $\odot$ is the element-wise multiplication, and $\tanh$ is the hyperbolic tangent function.
\begin{equation}
\begin{aligned}
    \mathbf{i}_t & =\sigma\left(\mathbf{W}_{i i} \mathbf{x}_t+\mathbf{b}_{i i}+\mathbf{W}_{h i} \mathbf{h}_{t-1}+\mathbf{b}_{h i}\right) \\
    \mathbf{f}_t & =\sigma\left(\mathbf{W}_{i f} \mathbf{x}_t+\mathbf{b}_{i f}+\mathbf{W}_{h f} \mathbf{h}_{t-1}+\mathbf{b}_{h f}\right) \\
    \mathbf{g}_t & =\tanh \left(\mathbf{W}_{i g} \mathbf{x}_t+\mathbf{b}_{i g}+\mathbf{W}_{h g} \mathbf{h}_{t-1}+\mathbf{b}_{h g}\right) \\
    \mathbf{o}_t & =\sigma\left(\mathbf{W}_{i o} \mathbf{x}_t+\mathbf{b}_{i o}+\mathbf{W}_{h o} \mathbf{h}_{t-1}+\mathbf{b}_{h o}\right) \\
    \mathbf{c}_t & =\mathbf{f}_t \odot \mathbf{c}_{t-1}+\mathbf{i}_t \odot \mathbf{g}_t \\
    \mathbf{h}_t & =\mathbf{o}_t \odot \tanh \left(\mathbf{c}_t\right)
\end{aligned}
\label{equ: lstm}
\end{equation}}

\modified{GCNConv~\cite{kipf2016semi} is a graph neural network module for learning spatial correlation in the data.
The feature for node $i$ at the $r$-th layer can be computed using the features of its neighbors $\mathcal{N}(i)$, including itself, by Equation~\ref{equ: gcnconv}:
\begin{equation}
    \mathbf{x}_i^{(r)}
    =
    \sum_{j \in \mathcal{N}(i) \cup\{i\}} 
    \frac{1}{\sqrt{D(i) \cdot D(j)}} 
    \cdot
    \left(\mathbf{W} \mathbf{x}_j^{(r-1)}\right),
    \label{equ: gcnconv}
\end{equation}
where $\mathbf{W}$ is the learnable parameters, $D(i)$ is the degree of node $i$ in the graph, and $\mathbf{x}_j^{r-1}$ is the feature of node $j$ at the previous layer.}

To provide a reasonable baseline model, the implementation of HiMSNet is kept as concise as possible.
The first step in HiMSNet is value embedding, i.e., the values and corresponding timestamps are embedded into a feature space using a linear layer, whose output dimension is 64.
Specifically, the missing values, which appear when a road is not monitored or no vehicle passes a monitored road, are replaced with learnable neural network parameters of equal size.
After the embedding, the drone modality has 2 layers of 1D convolution with kernel size 3 and stride 3 to down-sample the temporal resolution.
Then, each modality has an individual temporal encoder $\mathcal{H}_e$ consisting of 3-layer LSTM~\cite{hochreiter1997long} with hidden size 64, and their last step outputs will be used as the local feature $\mathbf{z}_p$.
The decoders for both segment-level and regional predictions ($\mathcal{H}_d$ and $\mathcal{H}_I$) are 2-layer MLPs with doubled hidden size (128) and output dimension 10 to predict 10 future time steps (30 minutes).
Contrary to modality-specific temporal encoders, the decoder is shared as the features have been combined.
The message encoding layer $\mathcal{R}_1$ is a 1 $\times$ 1 convolution layer that projects the features while keeping the dimension unchanged.
Accordingly, the message decoding layer $\mathcal{R}_2$ inherits the same structure.
The GME module is implemented with 3 layers of GCNConv~\cite{kipf2016semi}, each of them followed by ReLU activation and Layer Normalization~\cite{ba2016layernorm}.

\subsection{Dataset Creation}
\label{subsec: dataset creation}
Although existing drone datasets have already provided high-quality vehicle trajectory data, e.g., pNEUMA~\cite{barmpounakis2020new} and highD~\cite{highDdataset}, the aim of this work is to predict the traffic states for all road segments within a large urban region, which is beyond the scope of available datasets.
For this reason, detailed vehicle trajectory data are generated using the microscopic traffic simulator Aimsun and its road network for central Barcelona~\cite{casas2010traffic}.
\modified{Concretely, the dataset creation process consists of three steps: 1) \nameref{subsubsec: vehicle trajectories}, 2) \nameref{subsubsec: compute traffic variables} and 3) \nameref{subsubsec: sample extraction}.}

\subsubsection{Simulate Vehicle Trajectories}
\label{subsubsec: vehicle trajectories}
\begin{figure*}[!t]
    \centering
    \includegraphics[width=0.7\textwidth]{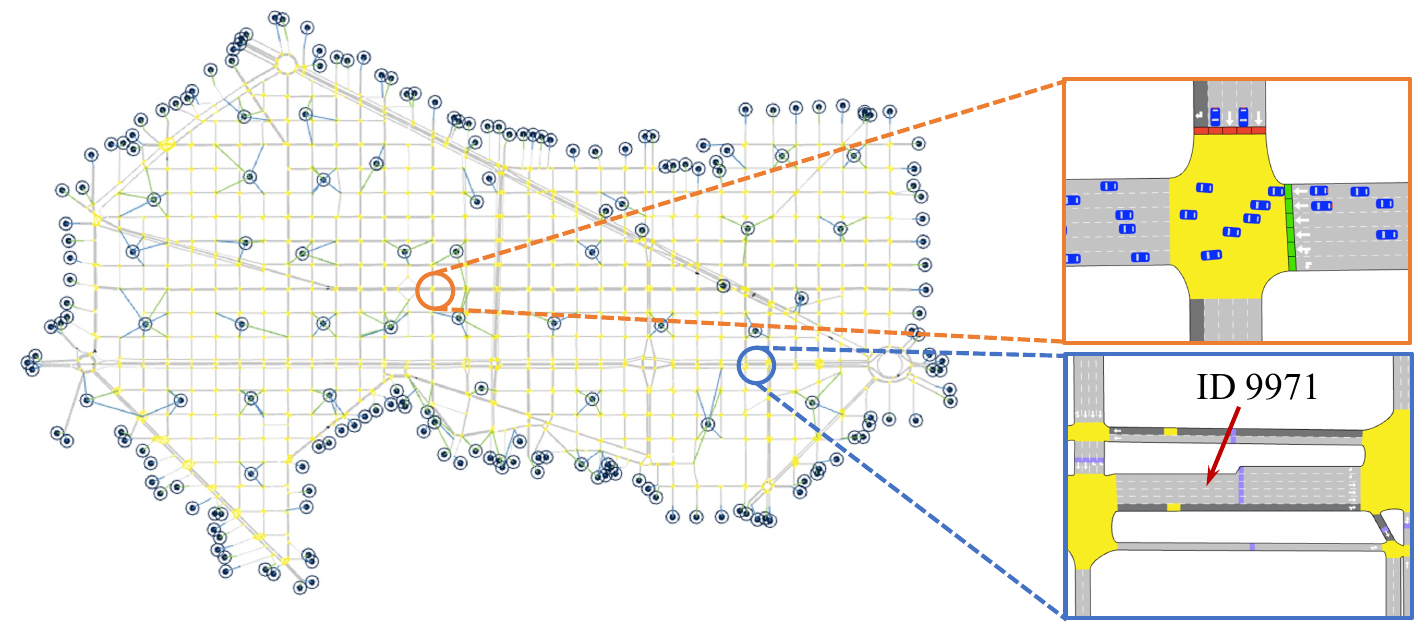}
    \caption{The simulated urban transportation network with detailed views of an intersection and a road segment (ID 9971), which will be used as visualization example in the experiments.}
    \label{fig: simulation network}
\end{figure*}

The simulation environment has been properly calibrated based on real-world traffic scenarios, and it has been applied in various analyses on urban traffic, such as network wide-perimeter control~\cite{tsitsokas2023two,kouvelas2017enhancing}.
As shown in Figure~\ref{fig: simulation network}, the network consists of many road segments, intersections as well as centroids ($\odot$ like) that serve as origins and destinations of vehicles. 
The fine details from microscopic traffic simulations allow great flexibility for subsequent processing and analysis, e.g., the computation of vehicle flow, density and speed.
Concretely, there are mainly three types of available information:
\begin{itemize}
    \item Network topology and the geometry of road segments.
    \item Vehicle locations at each simulation time step, referenced within road segments.
    \item The time when vehicles enter or exit a road segment.
\end{itemize}

In the simulation runs, the vehicles are created according to a specific OD matrix calibrated using historical data.
It is a square matrix with elements representing the number of vehicles going from an origin centroid to a destination in the time period of the demand.
However, the simulator provides only one such matrix, which can limit the diversity of the generated trajectory data and the ability of the machine learning model.
Therefore, the non-zero elements in the OD matrix are randomly augmented by the following operations, while the zero elements are kept unchanged:
\begin{itemize}
    \item randomly set an element to zero
    \item increase or decrease an element by a random percentage
    \item multiply all elements by a common random demand scale
\end{itemize}

\subsubsection{Compute Traffic Variables}
\label{subsubsec: compute traffic variables}
\modified{Using vehicle trajectory data from microscopic traffic simulation, this work computes point speed and segment speed to resemble the characteristics of inductive loop detectors and drones.}
Inductive loop detectors can only measure vehicle speeds at certain points, due to fixed installations.
Since vehicles can change their speeds within a road segment, the such speed measurements are valid only in its proximity and often cannot represent the situation of the whole segment.
On the contrary, drones can capture nearly all vehicles in their field of view, therefore it is possible to compute accurate segment speeds using all available vehicle trajectories.
\modified{In this work, all road segments are assumed to have a loop detector at their middle points. }

\begin{figure}[!t]
    \centering
    \includegraphics[width=0.48\textwidth]{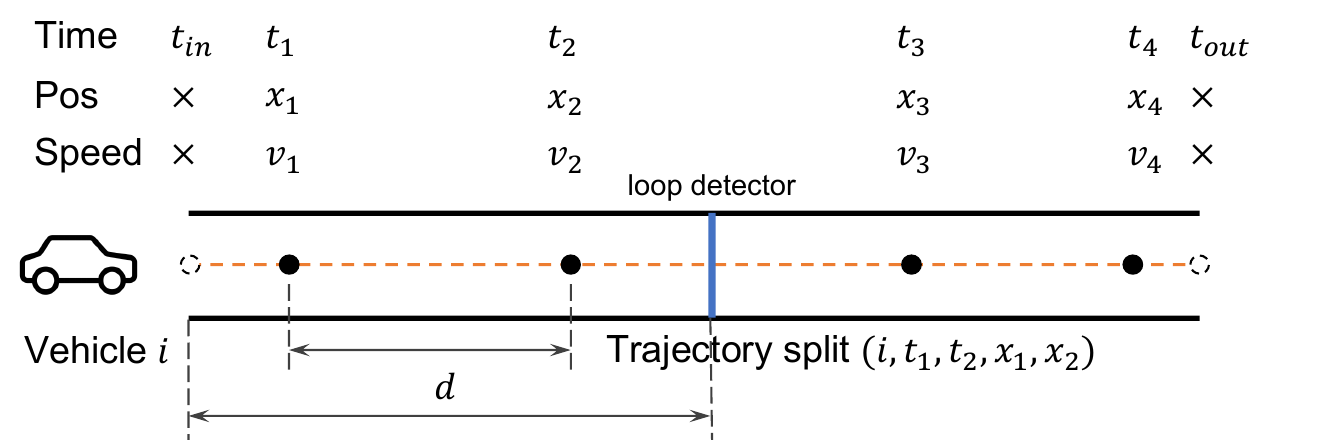}
    \caption{Traffic variable extraction with trajectory splits}
    \label{fig: trajectory processing}
\end{figure}

Figure~\ref{fig: trajectory processing} shows an example of a road segment along with the trajectory of one vehicle.
At each simulation time step ($t_1 \sim t_4$), the vehicle position and speed are known.
However, this work intends to develop a preprocessing pipeline also applicable to real-world drone datasets, and thus will avoid using the speed information and assume constant speed between the two known positions.
Because in reality, the vehicle speeds are not directly measured from drone videos, but are calculated using the vehicle locations at different frames.
Still, for the entry and exit points, only the timestamps ($t_{in}$ and $t_{out}$) are known, thus the positions have to be extrapolated from the closest known positions (e.g., the vehicle moves at speed $v_1$ from $t_{in}$ to $t_1$).

Then, two adjacent points can compose a \textit{trajectory split} $(i, t_s, t_e, x_s, x_e)$, which means vehicle $i$ travels from position $x_s$ to $x_e$ between time $t_s$ and $t_e$.
Afterward, the trajectory splits can be grouped by different time resolutions (e.g., every 5 s for drones and 3 minutes for loop detectors) to simulate the update frequency of drone data and loop detector data, or by spatial region to provide regional traffic data. 
For segment speed, we sum the travel distance $\Delta x = x_e - x_s$ and travel time $\Delta t = t_e - t_s$ of \textbf{all} trajectory splits in the road segment, and compute the segment speed following Edie's definitions of generalized flows and densities~\cite{edie1963discussion}: 
\begin{equation}
\label{equ: segment speed}
\bar{v}_s = \frac{\sum \Delta x}{\sum \Delta t}.
\end{equation}

\modified{For a loop detector installed $d$ meters from the start, if a vehicle crosses the detector's position during the time interval $(t_s, t_e)$, its corresponding position will satisfy $x_s < d < x_e$.
Then, the vehicle is \textit{detected} at the starting timestep of the trajectory split.
For example, the vehicle in Figure~\ref{fig: trajectory processing} is detected at $t_2$, because its trajectory crosses the loop detector between $t_2$ and $t_3$.}
We count the number of \textbf{detected} vehicles $N$ in a time interval (e.g., every 3 minutes), and arithmetically average their speeds to get point speed as
\begin{equation}
\label{equ: point speed}
\bar{v}_p = (\sum \frac{\Delta x}{\Delta t})/N.
\end{equation}

By choosing different time resolutions for Equation~\ref{equ: segment speed} and \ref{equ: point speed}, the simulated loop detector and drone measurements can be computed individually for each road segment.
Similarly, regional speed can be obtained by applying Equation~\ref{equ: segment speed} to all the trajectory splits in a region during a time interval. 

\subsubsection{Sample Extraction and Sensor Simulation}
\label{subsubsec: sample extraction}
\begin{figure*}[ht]
    \centering
    \includegraphics[width=0.7\textwidth]{"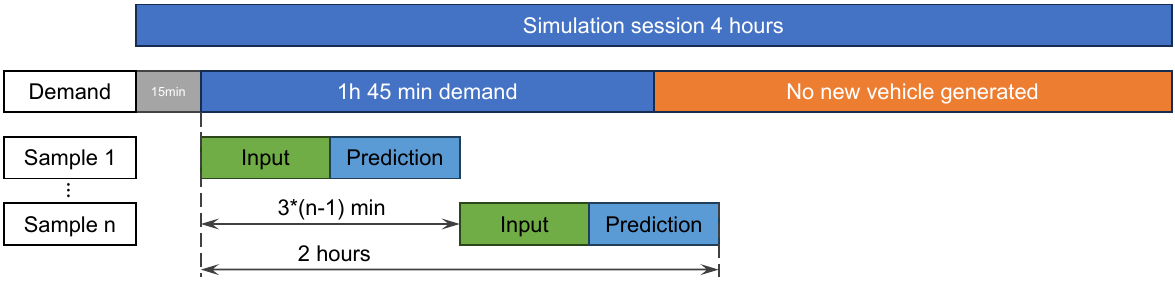"}
    \caption{Training sample extraction using a sliding window approach}
    \label{fig: sample extraction}
\end{figure*}

\modified{With the segment speed and point speed for each road segment, and average speed for each region,} the samples for the prediction model can be generated using a sliding window as in Figure~\ref{fig: sample extraction}.

Each simulation session lasts for 4 hours in the simulation world, with a 15-minute warm-up and 1 hour 45 min main demand, while the real-world run time differs according to the traffic demand.
This demand profile is intended to simulate the morning peak hours in the road network. 
The sliding window will start at the end of the warm-up period up to 2 hours after the beginning of the main demand, and the window size is 1 hour with a 3-minute step size between consecutive samples.
Inside a window, the first half is used as input and the second half is used as targeted output for the prediction model.
Therefore each simulation session will have 20 samples, each with 30-minute input and 30-minute output.

\begin{table}[h]
    \centering
    \caption{Input and target sequences in SimBarca}
    \label{tab: io sequences}
    \begin{tabular}{lccccc}
        \toprule
        \textbf{Speed} & \textbf{Sensor} & \textbf{Input}  & \textbf{In-Step}        & \textbf{Pred.} & \textbf{Pred-Step}  \\
        \midrule
        Segment             & Drone           & \checkmark  & 5 s        & \checkmark  & 3 min    \\
        Point               & LD   & \checkmark  & 3 min       & $\times$    & -         \\
        Regional            & -               & $\times$    & -           & \checkmark & 3 min     \\
        \bottomrule
    \end{tabular}
\end{table}

\modified{As shown in Table~\ref{tab: io sequences}, drones and loop detectors are assumed to provide segment speed and point speed respectively.
Point speed in the prediction window is excluded from targeted outputs because it can not represent the situation of the whole segment.
Meanwhile, regional speed is excluded from input sequences since it is highly-aggregated and can not be directly measured by an individual sensor. 
Therefore, SimBarca contains two input modalities, i.e., point speed and segment speed from simulated loop detectors and drone, and two prediction tasks, i.e., segment-level and regional speed prediction.}

In the input, loop detectors speeds are given every 3 minutes according to literature~\cite{loder2019understanding} and drones report speeds every 5 seconds.
Although in reality, drones are able to provide high-resolution video at 30 frames per second, the high data volume will create high demand for data transfer and computation in a real-time framework.
Therefore, this work simplifies the process with a lower temporal resolution while preserving accurate speed estimations from a traffic engineering perspective. 
The segment speeds computed by Equation~\ref{equ: segment speed} are used as \textbf{segment-level labels}, and the \textbf{regional labels} are obtained by applying the same formula to all trajectory splits within a region.
Both labels have 3 minute time step.

\section{Experiments}
\label{sec: experiments}

In this part, Section~\ref{subsec: dataset visualization} visualizes the dataset at both micro and macro levels, and Section~\ref{subsec: experiment settings} introduces the experiment settings for deep learning.
Section~\ref{subsec: overall prediction performance}, \ref{subsec: detailed analysis of prediction performance} and \ref{subsec: effects of hyperparameter} start the evaluation with a \textit{full-information} setting where the model has access to segment speeds and point speeds for all road segments.
Then, Section~\ref{subsec: partial observation and noisy data} and \ref{subsec: results with partial and noisy data} study a more realistic scenario with noisy data and partial sensor coverage.

\subsection{Visualization of Fine-grained and Aggregated Data}
\label{subsec: dataset visualization}
\begin{figure*}[!t]
    \centering
    \includegraphics[width=0.7\textwidth]{"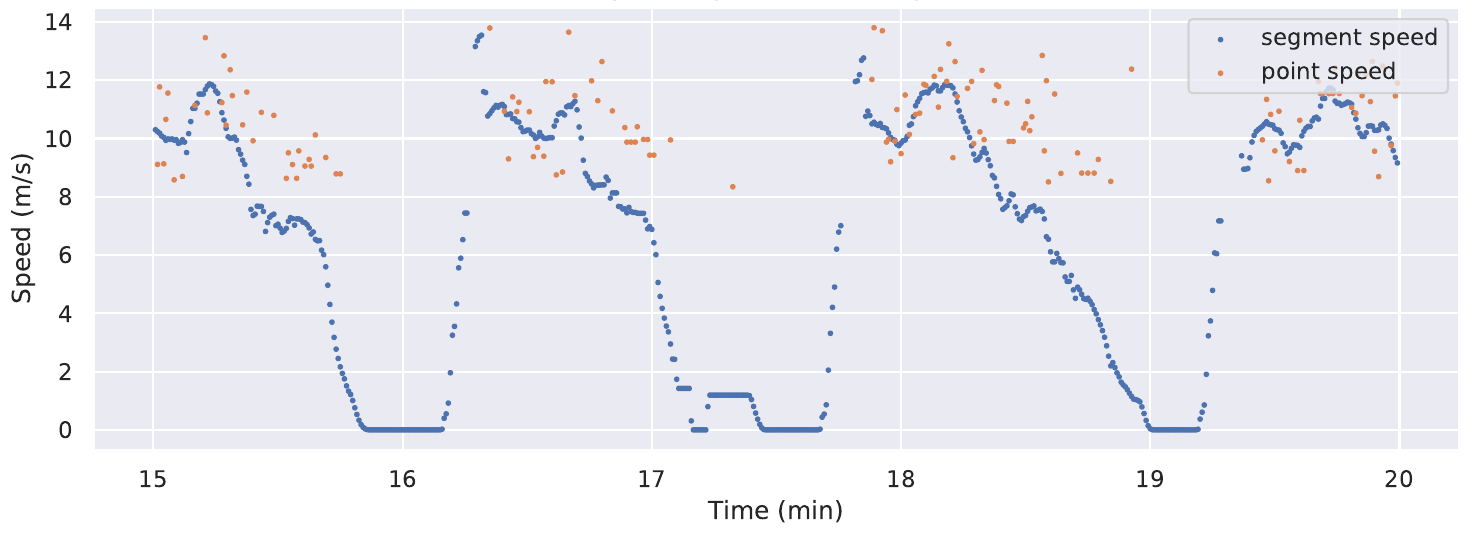"}
    \caption{Comparison of segment speed and point speed per simulation time step (0.5s), segment ID 9971. \modified{This visualization does not include missing values when a loop detector or a drone senses no vehicle.}
    }
    \label{fig: segment speed vs point speed}
\end{figure*}

\begin{figure*}[!t]
    \centering
    \includegraphics[width=0.7\textwidth]{"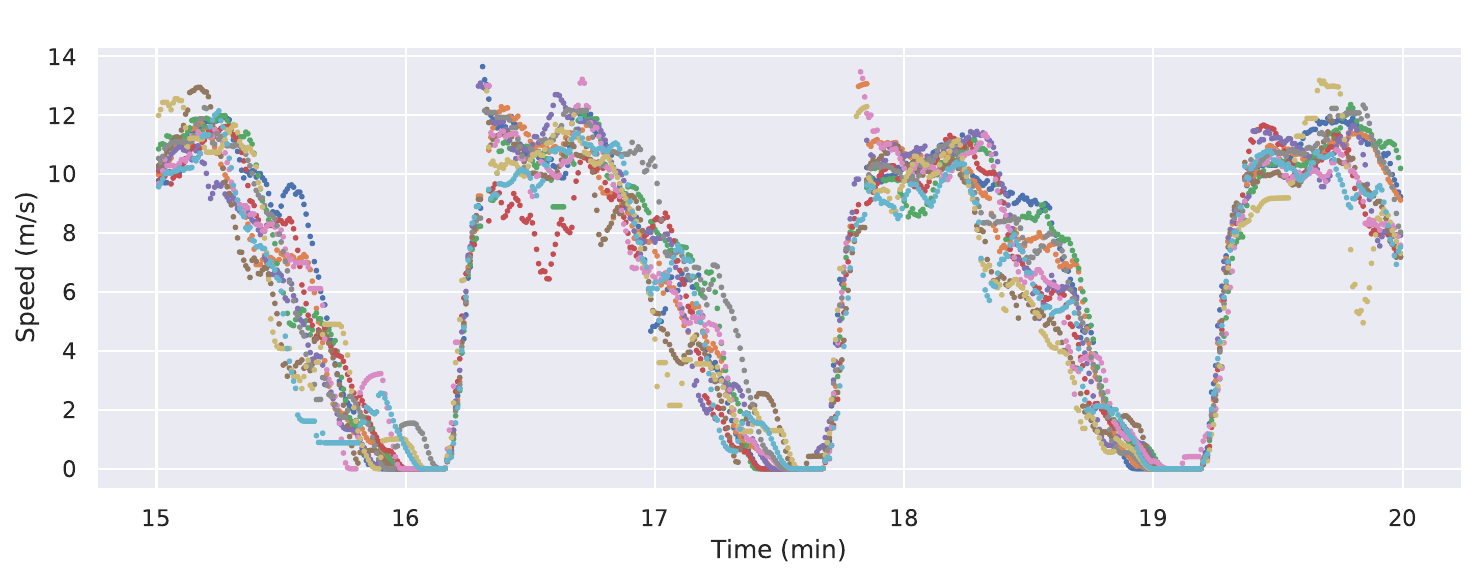"}
    \caption{segment speed from multiple simulation runs per simulation time step (0.5s), segment ID 9971. The time series are drawn as discrete points to reduce overlapping.}
    \label{fig: segment speed from multiple simulation runs}
\end{figure*}

Figure~\ref{fig: segment speed vs point speed} compares the segment speed and point speed per simulation time step, which are the finest possible details from the simulation data.
Traffic lights have caused a strong periodic pattern in segment speed, i.e., the acceleration and deceleration patterns in the segment speed.
On the other hand, the point speed can only capture the instant speed of vehicles at a specific location, and thus contains more stochastics due to randomized vehicle passing. 
From the aspect of congestion development and propagation, point speed does not provide as much information as the segment speed.
Besides, the randomized simulations have resulted in varying dynamics in the transportation network. 
For example, Figure~\ref{fig: segment speed from multiple simulation runs} shows the segment speed of the same road segment from multiple simulation runs.
The periodic pattern is consistent across different runs, but the details vary because of the randomized demand augmentation and vehicle generation processes.

\begin{figure*}[!t]
    \centering
    \subfloat[]{
        \includegraphics[height=0.3\textwidth]{"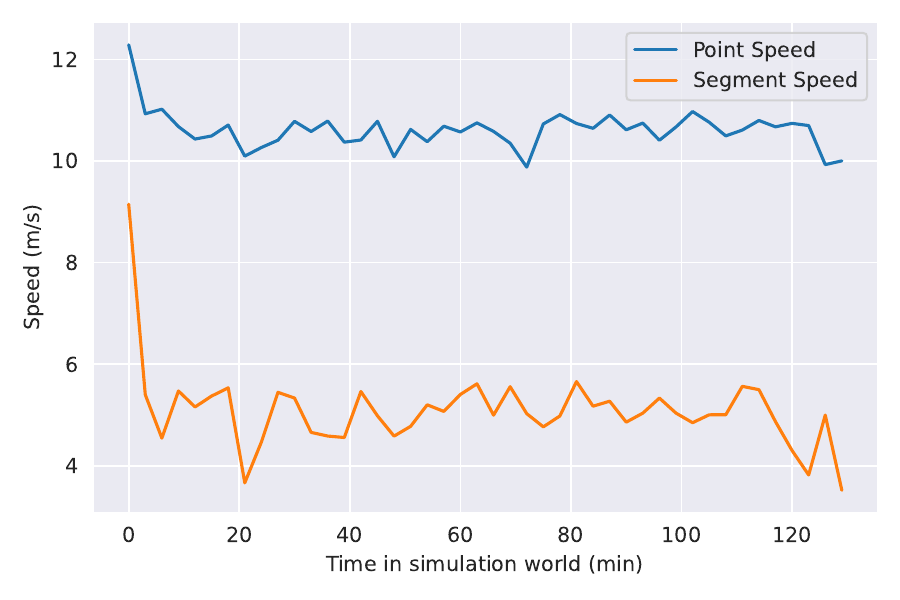"}
        \label{fig: speed comparison}
    }%
    \subfloat[]{
        \includegraphics[height=0.3\textwidth]{"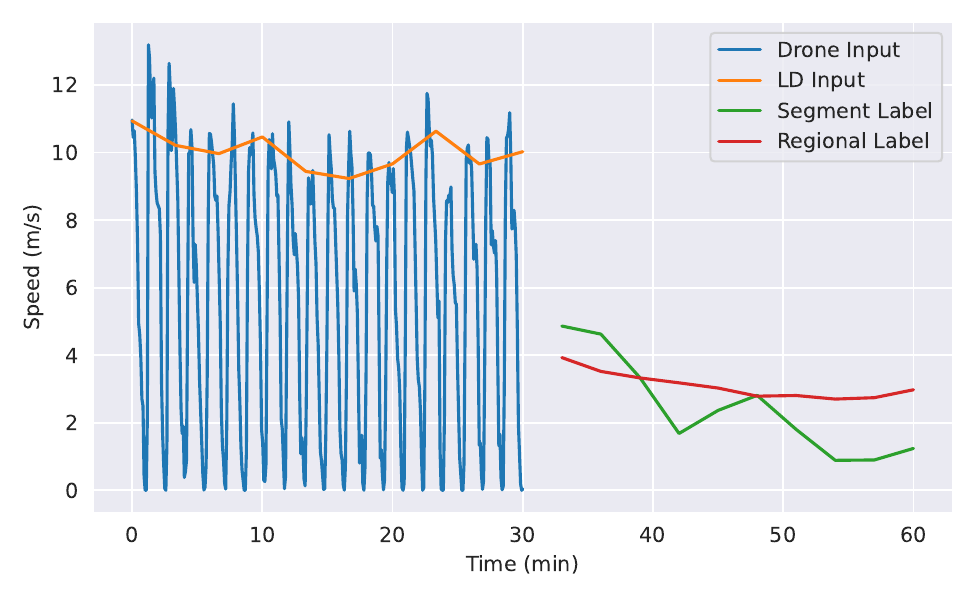"}
        \label{fig: data sample}
    }%
    \caption{Visualization of aggregated time series (a) and training sample (b) for road segment with ID 9971}
\end{figure*}

The fine-grained initial statistics are then aggregated to coarser time resolutions for the deep learning experiments.
The differences between point speed and segment speed are still clear in the aggregated data.
For example, Figure~\ref{fig: speed comparison} shows the segment speed and point speed for the road segment with ID 9971 (as marked in Figure~\ref{fig: simulation network}), both aggregated every 3 minutes.
\modified{In this example, point speed overestimates true segment speed roughly by 5 m/s, because a loop detector can not sense the vehicles stopping to wait for the red lights unless the queue is long enough to reach it}.

Figure~\ref{fig: data sample} shows an example training sample obtained by the sliding window approach.
In the input sequences (first 30 minutes), the segment speed (Drone Input) is given every 5 seconds, where the periodic pattern can be visually identified as the per-time-step case in Figure~\ref{fig: segment speed vs point speed}.
For example, the segment speed drops to zero periodically at every cycle of traffic signals due to the effect of the red phase.
The point speed (LD Input) is aggregated to a 3-minute interval as common practice, and still it overestimates the true speed, since the measured loop detector speed is influenced by the distance of the sensor from the stop line.
The second 30 minutes of the figure shows the prediction labels, including the future speed of the segment (Segment Label) and that of the region (Regional Label) this segment belongs to.
In this example, traffic congestion has already built up within the short time span of the training sample, as both the segment and regional speed have decreased.

\begin{figure}[!t]
    \centering
    \includegraphics[width=0.4\textwidth]{"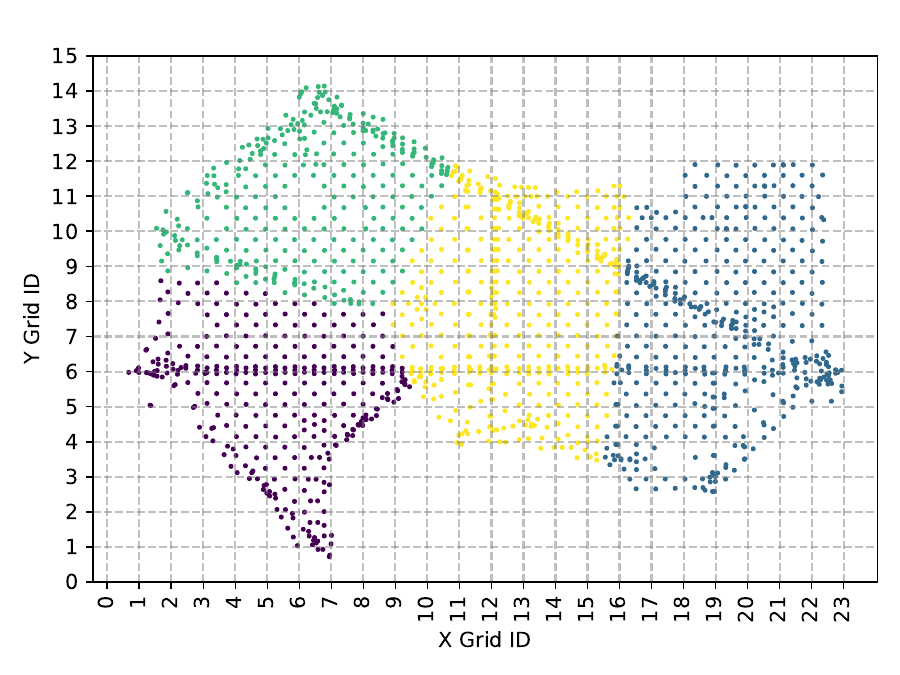"}
    \caption{Spatial regions and grids for drone monitoring.}
    \label{fig: region cluster and grid}
\end{figure}

In the regional prediction task, the road segments are clustered into 4 regions based on spatial distances using K-Means for simplicity.
More advanced clustering algorithms could also be utilized for this task to consider both road structure and historical traffic data~\cite{saeedmanesh2016clustering}.
In Figure~\ref{fig: region cluster and grid}, the center point of each road segment is marked with a point, and the color represents the region it belongs to.
The map is also discretized into grids with cell size $220~\times~220 m$, and it is assumed that a drone hovering at the center of a grid can monitor the traffic condition of all road segments, whose center points are inside the grid.

\begin{figure*}[!t]
    \centering
    \subfloat[]{\includegraphics[width=0.32\textwidth]{"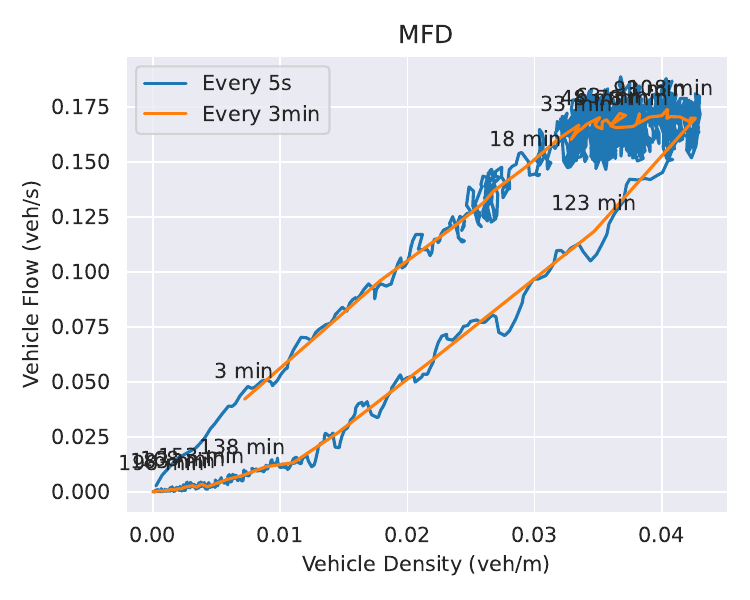"}}
    \subfloat[]{\includegraphics[width=0.32\textwidth]{"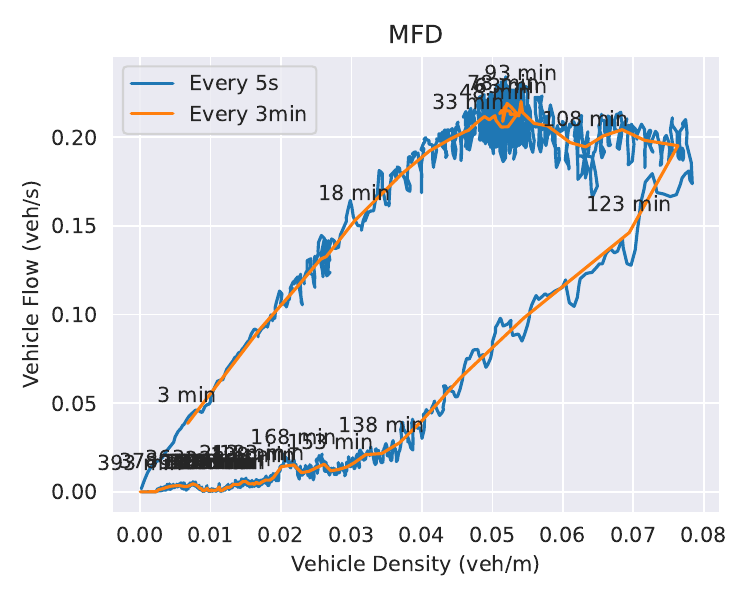"}}
    \subfloat[]{\includegraphics[width=0.32\textwidth]{"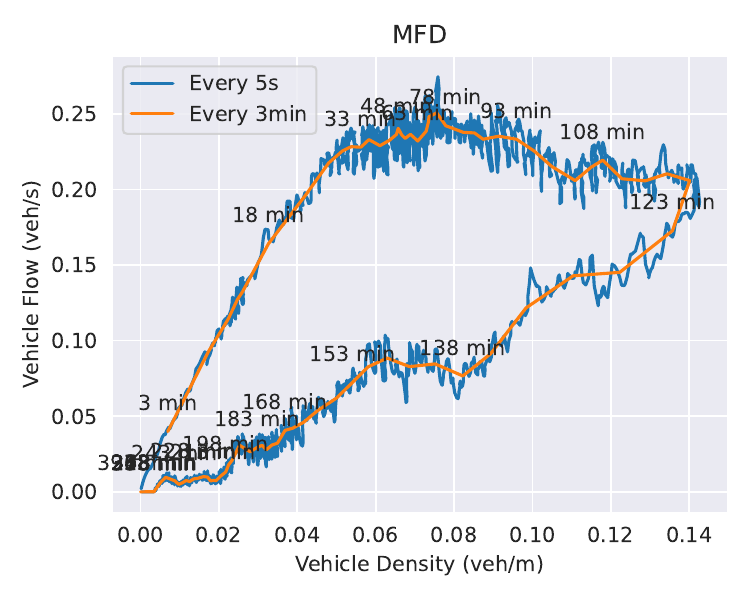"}}
    \caption{MFD under different demand scales, (a) 120\%, (b) 150\% and (c) 180\%. \modified{The points in the plots are pairs of vehicle flow and density values, linked according to simulation time indicated by the text annotations.
    These MFDs have hysteresis because after the duration with traffic demand, the simulation still runs to allow vehicles to clear the network. This exiting stage behaves differently due to the absence of new vehicles, and it is not included in sample extraction, as shown in Figure~\ref{fig: sample extraction}}.}
    \label{fig: mfd under different demand scales}
\end{figure*}

Figure~\ref{fig: mfd under different demand scales} shows the MFD of the whole road network under different demand scales, which is the augmentation factor multiplied with the demand matrix as mentioned in Section~\ref{subsubsec: vehicle trajectories}.
Notably, the exact number of vehicles in the simulation is still affected by the other two augmentations, and thus is not strictly proportional to the demand scale.
The vehicle flow is obtained by $q = D/tL$, where $D$ is the total distance traveled by all vehicles during the time interval of concern $t$, and $L$ is the total length of the road network.
Similarly, the vehicle density is obtained using $k = T/tL$ with $T$ being the total travel time of all vehicles.
Each subplot shows the MFD with $t$ being 5 seconds and 3 minutes, corresponding to the time resolution of drone and loop detector data respectively, and the data points are annotated with the time from the simulation start.

\begin{figure*}[!t]
    \centering
    \subfloat[]{\includegraphics[width=0.32\textwidth]{"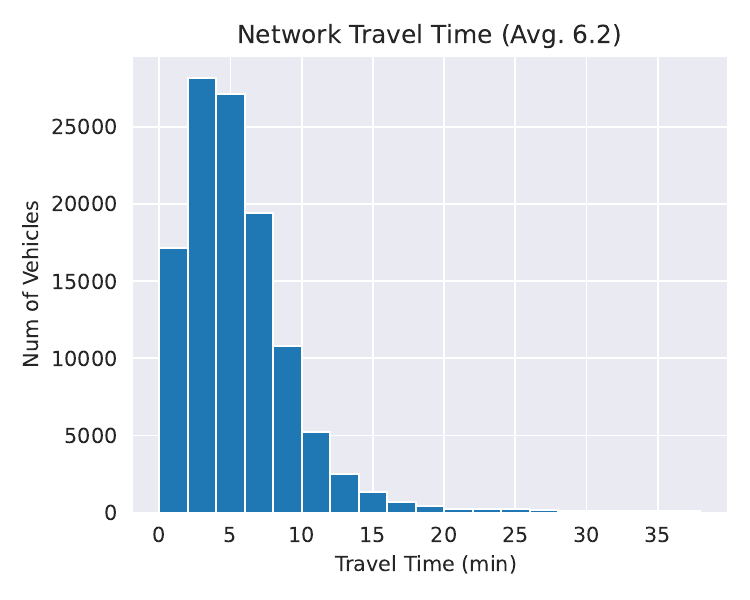"}}
    \subfloat[]{\includegraphics[width=0.32\textwidth]{"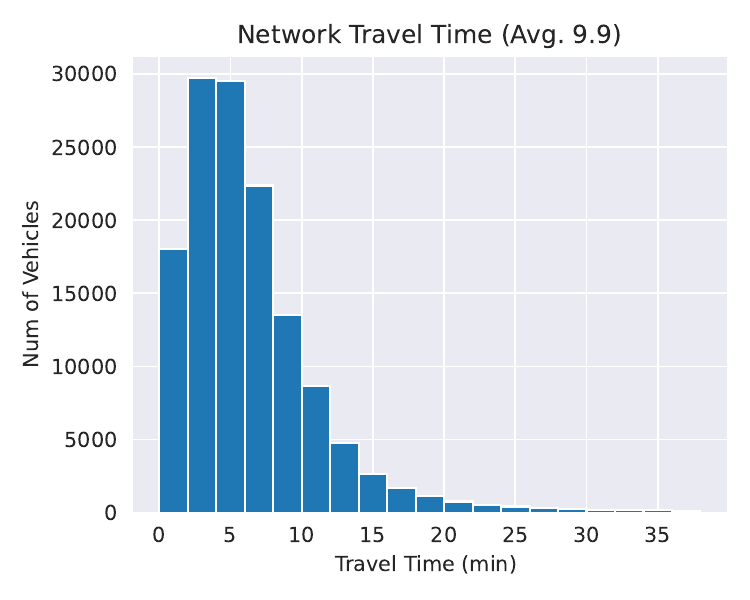"}}
    \subfloat[]{\includegraphics[width=0.32\textwidth]{"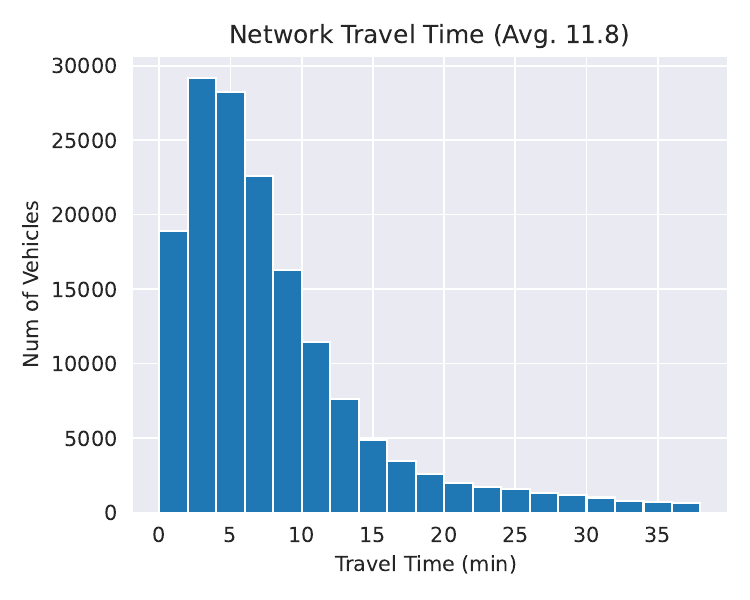"}}
    \caption{Network travel time distribution under different demand scales, (a) 120\%, (b) 150\% and (c) 180\%}
    \label{fig: travel time histogram under different demand scales}
\end{figure*}

These plots show that as the demand level increases, the vehicle flow becomes more saturated and the vehicle density reaches a higher maximum value.
Meanwhile, the network-level average speed also decreases greatly as the congestion becomes more severe.
Besides, the hysteresis loop is consistent with previous empirical and simulation observations~\cite{mahmassani2013urban}.
The increased congestion can be further evidenced by the travel time distribution of the network, as shown in Figure~\ref{fig: travel time histogram under different demand scales}. 
The increase in demand scale leads to a larger average and a longer tail in the travel time distribution, which means more and more vehicles are experiencing longer travel times.

\subsection{Experiment and Evaluation Settings}
\label{subsec: experiment settings}

SimBarca consists of 101 simulation runs, including one with the original demand profile unmodified, \modified{provided along side the simulation network by Aimsun~\cite{casas2010traffic}}, and the other 100 containing different demands augmented from the original one.
The training set contains 75 randomly selected simulation runs, and the rest are for testing.
The data graph is undirected and contains 1570 nodes (road segments) and 2803 edges.

By default, HiMSNet is trained for 30 epochs using Adam optimizer with a learning rate 0.001, weight decay $1 \times 10^{-4}$, $\beta = (0.9, 0.999)$ and batch size 8.
The learning rate scheduler gradually increases the learning rate from 0 to the chosen value during the first epoch, and decreases it to 10\% and 1\% of the chosen value at 70\% and 85\% of the total epochs. 
All implementations are based on PyTorch, and the experiments are done on a computer with an Nvidia RTX 4090 GPU. 

Following the common practice in traffic forecasting literature~\cite{li2018dcrnn}, the prediction results are evaluated with three metrics: MAE, Root Mean Square Error (RMSE) and Mean Absolute Percentage Error (MAPE).
Their vector forms for a pair of prediction ($\hat{\mathbf{y}}$) and ground truth ($\mathbf{y}$) are defined as follows:

\begin{equation}
    \label{equ: mean prediction evaluation metrics}
    \begin{cases}
        \text{MAE}  & =  \frac{1}{n}\sum_i^n | \hat{\mathbf{y}}_i - \mathbf{y}_i| \\
        \text{RMSE} & =  \sqrt{\frac{1}{n} \sum_i^n (\hat{\mathbf{y}}_i - \mathbf{y}_i)^2} \\ 
        \text{MAPE} & =  \frac{1}{n} \sum_i^n |(\hat{\mathbf{y}}_i - \mathbf{y}_i)/\mathbf{y}_i| 
    \end{cases},
\end{equation}
where $n$ is an index for flattened predictions and labels.

In the evaluation, the speed values take the unit of m/s, and the metrics are calculated for each road segment (or region) and then averaged over all segments (regions).
Since a zero segment speed in the label will result in infinite MAPE, we only evaluate MAPE when the speed value is greater than 1~m/s, and we refer to this modified metric as MAPE*.
These metrics are reported for 15 min and 30 min prediction windows for both segment-level and regional task, which correspond to time steps 5 and 10 in the dataset.

\subsection{Overall Prediction Performance}
\label{subsec: overall prediction performance}

Table~\ref{tab: comparison with baseline methods} compares the prediction errors of HiMSNet with three simple baselines \modified{to demonstrate the necessity of drone data and the effect of HiMSNet modules}, best results in \textbf{bold}.

Last Observation (LO) uses the last observed speed as the segment-level prediction, while Input Average (IA) uses the average speed of the input sequence.
Both of them can be configured to use loop detector (ld) or drone (drone) data, and the regional prediction is made by averaging the segment-level predictions.
The Label Average (LA) baseline uses the average speed of the whole test set labels as prediction, which applies to both prediction tasks.
In fact, the LA baseline is the best possible constant-value prediction that leads to the lowest MAE. 
The second group of methods are variants of HiMSNet, where we present a brief ablation study to examine the importance of different components.
The exact settings are: \modified{replacing the lightweight GCNConv~\cite{kipf2016semi} with a Transformer Encoder (TF)~\cite{vaswani2017attention}}, disable the GME module (/gnn) to exclude vicinity information, only use loop detector modality in the input (ld), and only use the drone modality (drone) in the input.

The IA and LO baseline with loop detector data have significantly higher errors than the variants using drone data, which originates from the difference between loop detector measurements and the drone measurements as shown in Figure~\ref{fig: segment speed vs point speed}.
IA(drone) also outperforms LO(drone), which is expected as the drone-measured speed has a high update frequency (every 5s) to reveal the traffic dynamics within a signal cycle. 
Therefore using the last observation as a prediction will be very biased towards the end of the cycle.
The LA baseline has the lowest error among the first group of results, and HiMSNet outperforms all the baselines, especially in the regional prediction task.
\modified{Transformer~\cite{vaswani2017attention} is generally considered as a more advanced architecture, however, since the HiMSNet(TF), has very similar errors as the HiMSNet with GCNConv~\cite{kipf2016semi}, we default to the simpler one.}

\begin{table*}[htbp]
    \centering
    \caption{Prediction errors for road segment task and regional task in perfect information case}
    \resizebox{\textwidth}{!}{%
    \begin{tabular}{lcrccrc|crccrc}
        \toprule
        \multirow[c]{2}{*}{Model}& \multicolumn{3}{c}{Segment 15 min} & \multicolumn{3}{c|}{Segment 30 min} & \multicolumn{3}{c}{Regional 15 min} & \multicolumn{3}{c}{Regional 30 min} \\
        \cmidrule(lr){2-4} \cmidrule(lr){5-7} \cmidrule(lr){8-10} \cmidrule(lr){11-13}
                      & MAE  & MAPE*    & RMSE & MAE  & MAPE*    & RMSE & MAE  & MAPE*    & RMSE & MAE  & MAPE*    & RMSE \\
        \midrule
        IA(ld)        & 4.50 & 142.7\% & 5.33 & 4.58 & 144.2\% & 5.45 & 6.02 & 174.7\% & 6.07 & 6.18 & 192.8\% & 6.27 \\
        IA(drone)     & 1.53 & 42.3\%  & 2.06 & 1.70 & 45.7\%  & 2.32 & 2.38 & 72.4\%  & 2.51 & 2.54 & 83.9\%  & 2.75 \\
        LO(ld)        & 4.47 & 141.6\% & 5.32 & 4.55 & 143.3\% & 5.44 & 5.95 & 172.2\% & 5.99 & 6.11 & 190.5\% & 6.02 \\
        LO(drone)     & 3.26 & 90.5\%  & 4.23 & 3.33 & 92.4\%  & 4.34 & 2.70 & 80.4\%  & 2.78 & 2.87 & 92.3\%  & 3.01 \\
        LA            & 1.24 & 28.4\%  & 1.90 & 1.43 & 31.7\%  & 2.18 & 0.83 & 26.8\%  & 1.08 & 1.01 & 34.3\%  & 1.34 \\
        \cmidrule(lr){1-13}
        HiMSNet       & 1.01 & \textbf{23.2\%} & \textbf{1.68} & 1.17 & 26.3\% & 1.94 & 0.20 & 6.1\% & 0.28 & 0.30 & 9.9\% & 0.46 \\ 
        \modified{HiMSNet(TF)}   & \textbf{1.00} & \textbf{23.2\%} & 1.69 & \textbf{1.15} & \textbf{26.1\%} & \textbf{1.93} & \textbf{0.19} & 6.1\% & 0.28 & \textbf{0.29} & 9.8\% & 0.46 \\
        HiMSNet/gnn   & 1.03 & 24.0\% & 1.74 & 1.21 & 27.4\% & 2.03 & 0.20 & 6.2\% & 0.28 & 0.31 & 10.5\%  & 0.47 \\
        HiMSNet(drone)& \textbf{1.00} & \textbf{23.2\%} & \textbf{1.68} & 1.16 & 26.3\% & \textbf{1.93} & \textbf{0.19} & \textbf{6.0\%} & \textbf{0.27} & \textbf{0.29} & \textbf{9.6\%} & \textbf{0.45} \\ 
        HiMSNet(ld)   & 1.13 & 26.3\% & 1.84 & 1.29 & 29.3\% & 2.10 & 0.55 & 15.5\% & 0.63 & 0.75 & 22.2\% & 0.88 \\
        \bottomrule
    \end{tabular}%
    }
    \label{tab: comparison with baseline methods}%
\end{table*}%

For the regional task, most of the HiMSNet variants have close prediction performance, except that HiMSNet(ld) clearly has higher errors than the other variants.
Moreover, although the drone modality has been excluded from the input of HiMSNet(ld), the training labels of this model are still computed using the vehicle trajectory data, and in reality, such data are not possible with loop detectors only.
Compared to using drone data directly as input, HiMSNet(ld) faces additional challenges because of the difference between point speeds and segment speed (Figure~\ref{fig: segment speed vs point speed}). 
This performance difference suggests that the translation from point speed to segment speed can not be easily learned, and also highlights the necessity of drone data.

Similarly, in the segment-level prediction task, the model using only drone inputs HiMSNet(drone) has the lowest errors among all the HiMSNet variants, and HiMSNet(ld) has the highest.
This observation again highlights the importance of the drone modality, and also verifies the difficulty of learning the translation from point speed to segment speed.
Under the \textit{full-information} setting, the single-modality HiMSNet(drone) even slightly outperforms the multi-modality HiMSNet, suggesting that the loop detector modality may be uninformative when the same locations are covered by drones.
Different from the regional task, the inclusion of vicinity features shows higher impacts on the segment-level task, as removing the GME module in HiMSNet(/gnn) results in higher prediction errors, compared to the default HiMSNet.

\subsection{Detailed Analysis of Prediction Performance}
\label{subsec: detailed analysis of prediction performance}

\modified{This section visualize the model's predictions, as well as the distribution of prediction errors by spatial locations, demand scales and average speeds.} 

\begin{figure*}[ht]
    \centering
    \includegraphics[width=\textwidth]{"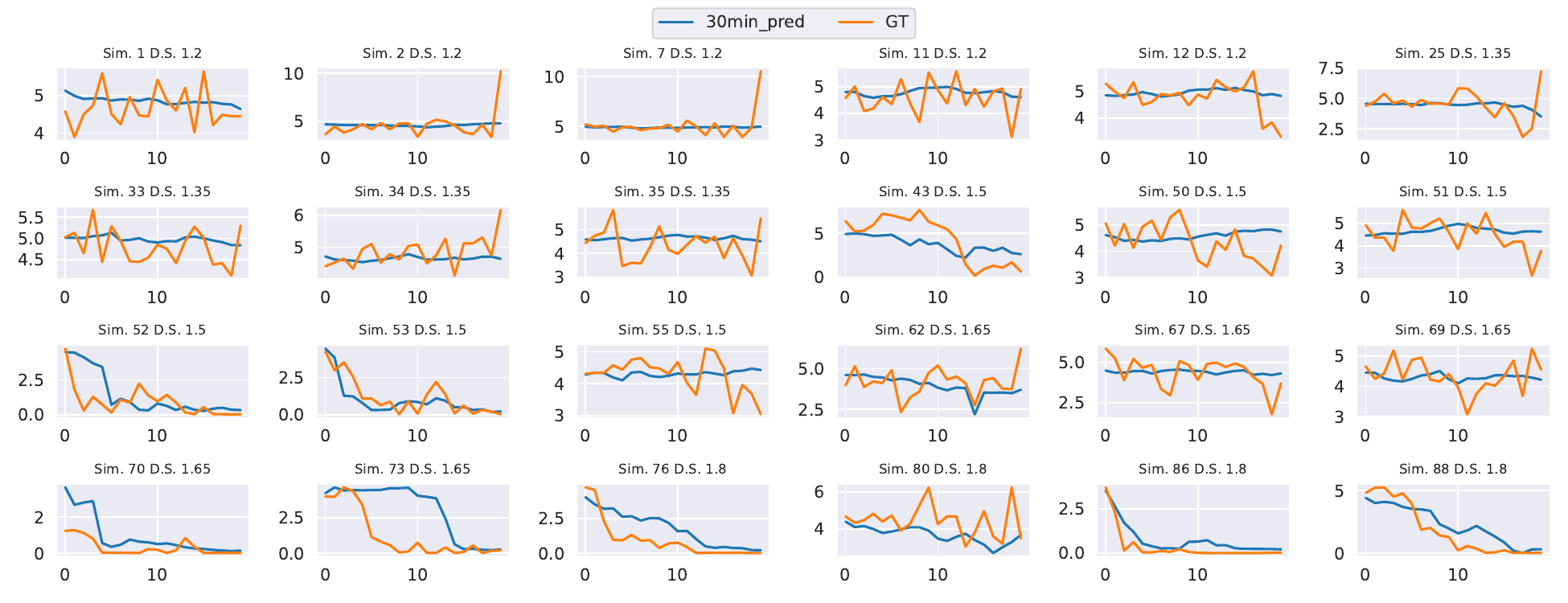"}
    \caption{30-min-ahead prediction for a road segment (ID 9971) at different simulation sessions. In each subplot, the $x$-axis is the time step index (3 min per step), and $y$-axis is the speed value in m/s. The title indicates the number of the simulation session (Sim.) out of the total 101 sessions, along with the demand scale (D.S.) of the session.}
    \label{fig: 30-min-ahead prediction 9971}
\end{figure*}

Traffic management and control often require predictions ahead of time to allow timely actions, and therefore the 30-min-ahead prediction is of particular interest.
Figure~\ref{fig: 30-min-ahead prediction 9971} shows the 30-min-ahead prediction of HiMSNet for one road segment with the groundtruth, where each subplot corresponds to a different simulation session.
As shown in the plot, the groundtruth can often have rapid changes over time, which reflects the dynamic nature of urban traffic flow. 
Due to different traffic demands, the speed of a segment also changes vastly from one simulation run to another.
For example, in Session 7 of Figure~\ref{fig: 30-min-ahead prediction 9971}, the segment speed is relatively stable (mostly around 5 m/s), but in Session 86, the speed decreases from 2.5 down to 0 because of heavy congestion.
Since the time step is 3 minutes, the prediction horizon of 30 minutes corresponds to 10 steps.
The results show that HiMSNet can capture the general trend of the speed changes (Session 11), but at the same time, it has difficulties predicting the rapid changes in the segment speed.
The model can predict congestion well in advance (Session 53), but there are also cases where HiMSNet only knows the congestion after it happens (Session 73).

\begin{figure}[!t]
    \centering
    \includegraphics[width=0.4\textwidth]{"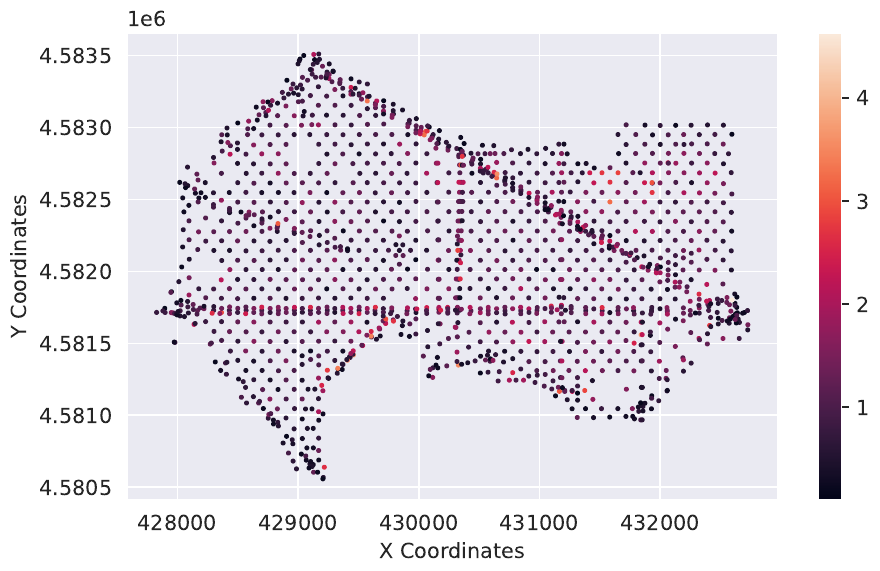"}
    \caption{Average test set MAE (unit m/s) for different locations, coordinate format EPSG:32601}
    \label{fig: mae by place}
\end{figure}

Figure~\ref{fig: mae by place} illustrates the spatial distribution of the prediction MAE of HiMSNet on SimBarca test set.
This MAE is averaged over all the prediction horizons, which indicates the overall performance.
Most of the road segments have low prediction MAE, but there are still a few places with higher errors, which suggests the traffic states there are more difficult to predict.
From the perspective of traffic management, these locations are probably worth close monitoring.

\begin{figure*}[!t]
    \centering
    \subfloat[]{
        \includegraphics[width=0.32\textwidth]{"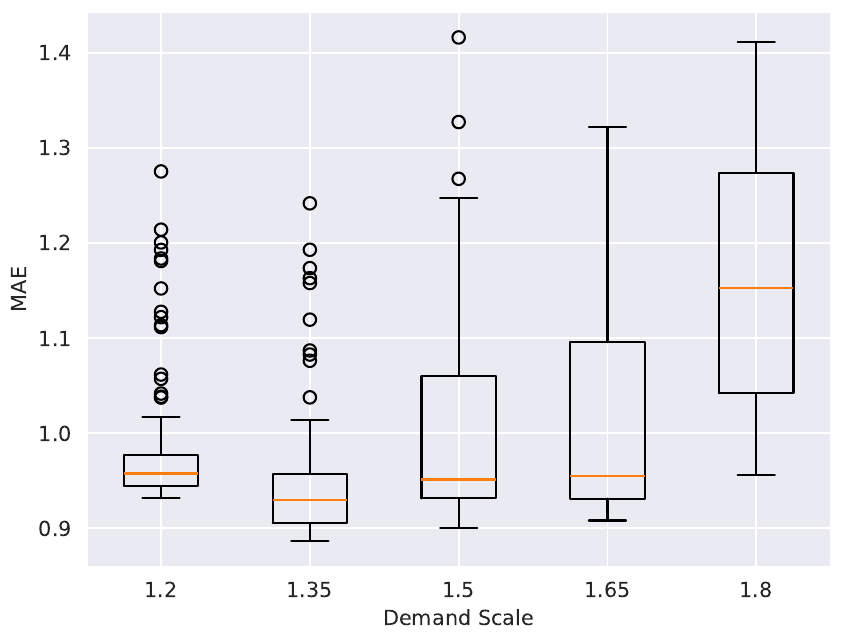"}
        \label{fig: segment mae by demand scale}
    }
    \subfloat[]{
        \includegraphics[width=0.32\textwidth]{"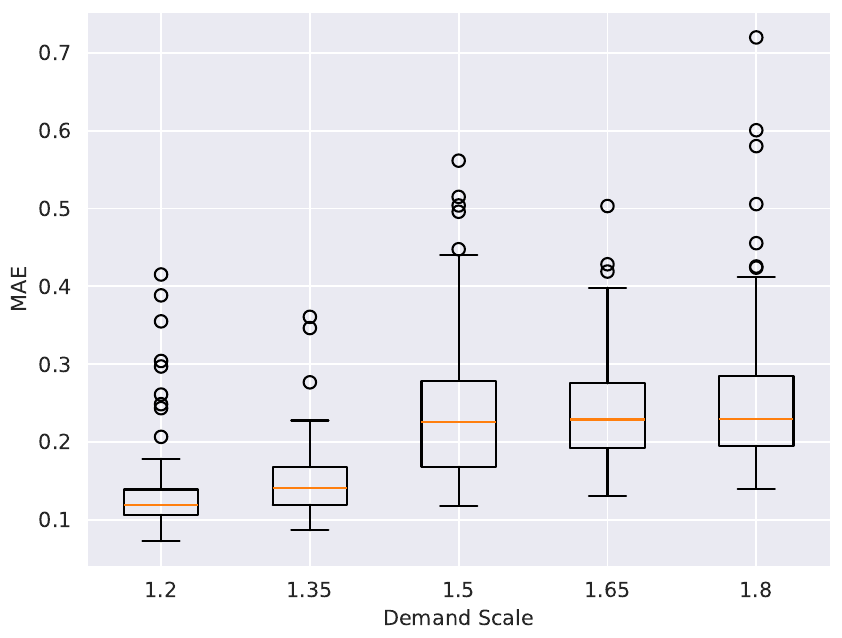"}
        \label{fig: regional mae by demand scale}
    }
    \subfloat[]{
        \includegraphics[width=0.32\textwidth]{"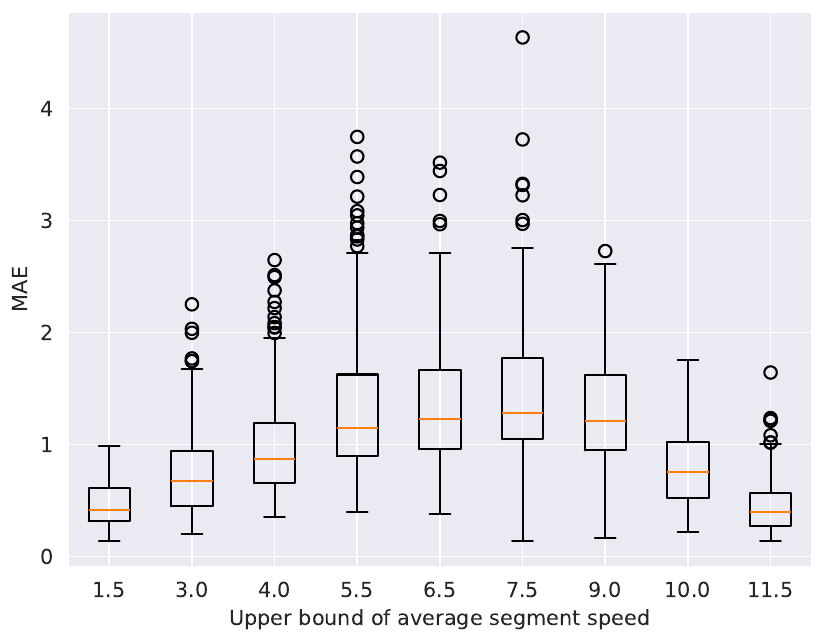"}
        \label{fig: segment mae by avg speed}
    }
\caption{Average MAE (unit m/s) over all prediction horizons grouped by different factors. (a) by demand scale for the segment-level task, (b) by demand scale for the regional task, (c) by average segment speed in the test set (segment-level task).}
\label{fig: mae by different factors}
\end{figure*}

To further analyze the error distribution and identify the challenging scenarios in urban traffic forecasting, Figure~\ref{fig: mae by different factors} shows the box plots for the average MAE over all prediction horizons grouped by different factors.
First of all, the prediction error increases as the demand scale increases, and a scale at 180\% will lead to significantly higher errors. 
This trend is evident in both segment-level (Figure~\ref{fig: segment mae by demand scale}) and regional tasks (Figure~\ref{fig: regional mae by demand scale}).
This is consistent with the intuition that a higher traffic demand will lead to heavier congestion and thus will induce more dynamics in the traffic flow, making the prediction more challenging.
Figure~\ref{fig: segment mae by avg speed} shows that the road segments with intermediate average speeds are harder to predict, while the ones with very low or very high average speeds have much lower errors.
This is probably because predicting the speed of a road segment will be easy when the road segment is always congested or always free-flowing.
However, when it often has a changing status (and thus intermediate average speed), accurate speed predictions will require the model to punctually capture the dynamics in the network traffic flow, which is more challenging.

\subsection{Effects of Hyperparameters}
\label{subsec: effects of hyperparameter}

\begin{figure*}[!t]
    \centering
    \subfloat[]{
        \includegraphics[width=0.35\textwidth]{"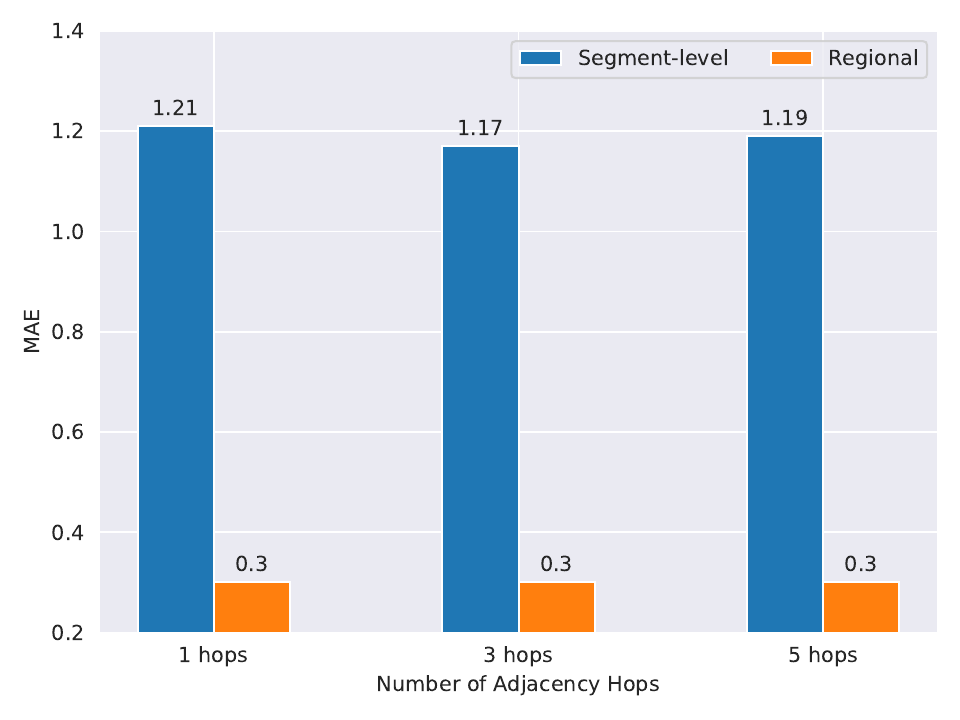"}
        \label{fig: 30 min mae by adj hop}
    }
    \subfloat[]{
        \includegraphics[width=0.35\textwidth]{"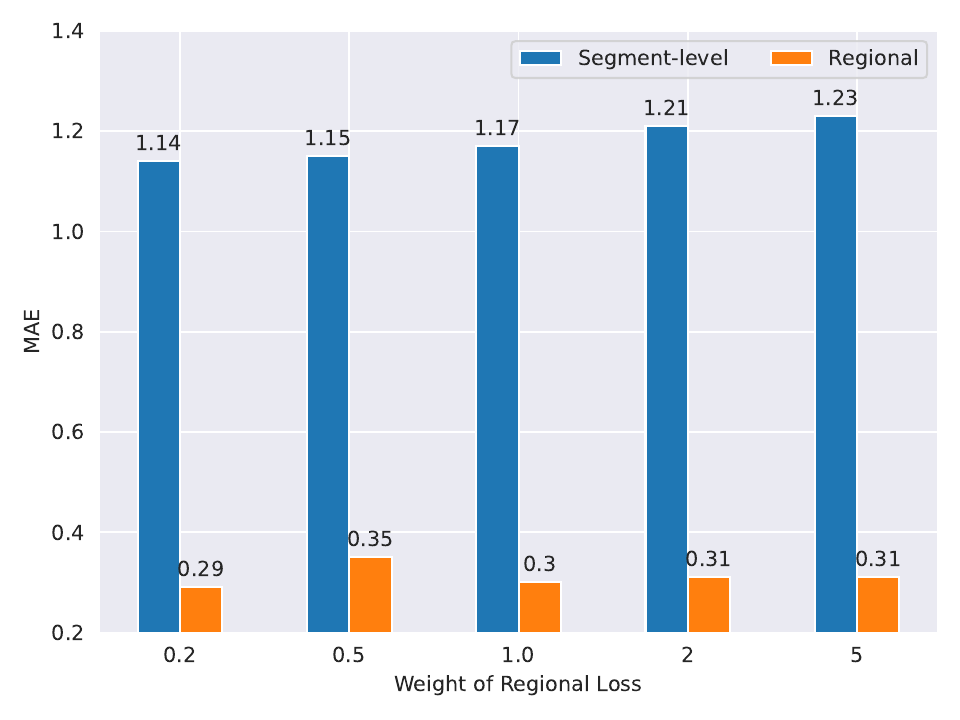"}
        \label{fig: 30 min mae by loss weight}
    }
    \\
    \subfloat[]{
        \includegraphics[width=0.35\textwidth]{"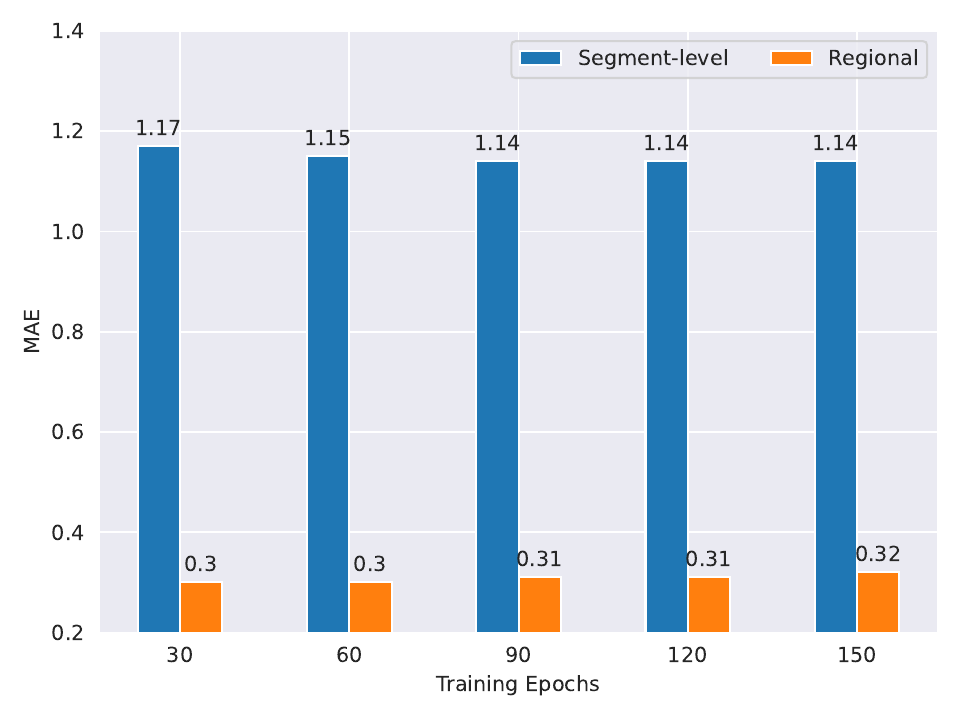"}
        \label{fig: 30 min mae by epochs}
    }
    \subfloat[]{
        \includegraphics[width=0.35\textwidth]{"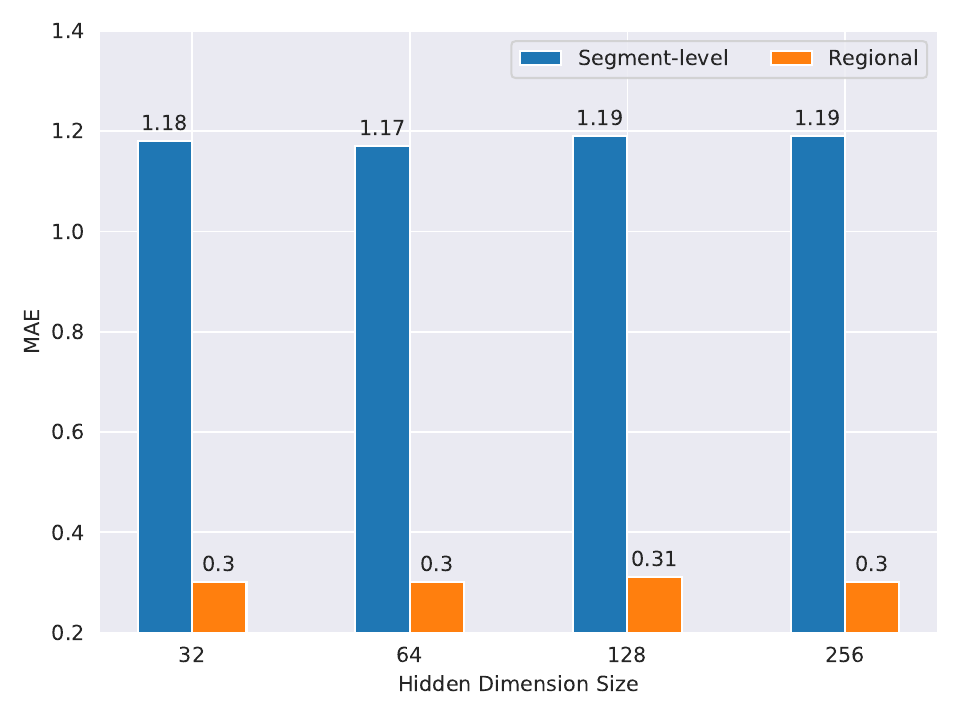"}
        \label{fig: 30 min mae by hidden dim}
    }
\caption{30-min-ahead prediction MAE (unit m/s) under different experiment settings. (a) by adjacency hop, (b) by loss weights, (c) by the number of training epochs and (d) by the hidden dimension.}
\label{fig: hyperparameter comparison}
\end{figure*}

\modified{This section compares the performance of the model under different hyper-parameters to justify the choice of default training settings}.
Figure~\ref{fig: hyperparameter comparison} compares the 30-min-ahead prediction MAE under different hyperparameter settings, including adjacency hops, loss balance between tasks, training epochs and model sizes.
The ablation study in Table~\ref{tab: performance with noise and partial observation} has shown that vicinity information is important for the segment-level prediction task, which highlights the effectiveness of the GME module.
Still, the scope of the considered neighborhood should be properly chosen.
In Figure~\ref{fig: 30 min mae by adj hop}, a single GCNConv~\cite{kipf2016semi} layer with 1-hop adjacency will strictly follow the road network definition and only take into account the immediate neighbors of a road segment.
Although this message passing scheme is stacked 3 times in the GME module, it still turns out to be suboptimal due to the limited adjacency scope.
On the contrary, a layer with 5-hop adjacency will have a large scope and include the nodes that are 5 hops away from a road segment, and thus it may introduce much irrelevant information.
Therefore in the three experimented setups, the 3-hop adjacency layer achieves the best performance for the segment-level task, as it has the most suitable vicinity scope.

Figure~\ref{fig: 30 min mae by loss weight} presents the regional and segment-level prediction errors when the weight of the regional task is set to different values, and the weight of the segment-level task is always fixed at 1.
When the regional task has a low loss weight, the model will be able to focus more on the segment-level task and thus achieve lower errors there.
However, the trend in the regional task is unclear when it is either underweighted or overweighted, and therefore in all other experiments, both tasks are chosen to have equal weight of 1 for simplicity.
Figure~\ref{fig: 30 min mae by epochs} compares the prediction errors at different training epochs, and a longer training duration will generally lead to lower error for the segment-level task but the regional task will be compromised.
Therefore in all other experiments, a duration with 30 epochs is chosen for both simplicity and a balance of the two tasks.
Figure~\ref{fig: 30 min mae by hidden dim} shows the prediction errors when the model size varies, where a larger model size can not lead to better performance.
For this reason, a default model size of 64 has been chosen for the model.

\subsection{Experiment Settings with Partial and Noisy Data}
\label{subsec: partial observation and noisy data}

In previous experiments, the model has access to the speed data of all road segments, and all data are free from noise.
However, a city can not install loop detectors on every road segment, or fly drones to cover the whole city all the time, and measurements from these sensors are often subject to noise.
Therefore, this section investigates a more realistic scenario with partial observations and noisy data, abbreviated as PN.

For partial observations, the loop detector modality has 10\% coverage, i.e., 157 road segments out of the 1570 segments are equipped with loop detectors. 
The segments with loop detectors are randomly selected after filtering out the locations with more than 10\% invalid values.
The same set of loop detector locations are applied to all simulation runs (both training and testing), since they are installed infrastructure and can not be changed easily.
As for drones, it is assumed that each drone can cover a grid cell of $220~\times~220 m$ while following common regulations on flying height, and 21 drones (also 10\%) are randomly placed over the 212 non-empty cells in Figure~\ref{fig: region cluster and grid}.
Contrary to loop detectors, drones are flexible monitoring agents, and they are randomly relocated every 3 minutes, without considering the time to move between cells.

Multiplicative Gaussian noise is added to the segment-level speed data.
The noises have zero mean and standard deviation 0.05 and 0.15 for loop detector and drone modality.
Notably, partial observation and noisy data are introduced in the training and evaluation phase in different ways.
In the training phase, both the inputs and segment-level labels are partial and noisy.
Meanwhile, the regional labels are computed with partial segment-level details, i.e., total vehicle travel distance and time in observed road segments.
During evaluation, the input data remain partial and noisy, but the labels are fully visible and noise-free. 
This approach is intended to assess the model's ability to infer the traffic dynamics of the entire road network even with access to only partial and noisy data.

In all the experiments, HiMSNet is trained to predict the future segment speed and regional speed, which are not available from loop detectors.
Therefore, the models HiMSNet(ld) and HiMSNet\_PN(ld) have also benefited from the drone data in the training phase, despite that their inputs have loop detector data only.
In fact, they are trained to predict future drone-measured traffic states by using past loop detector data only.
\modified{This corresponds to a more realistic scenario where both drone and loop detector data are collected for training, but drone data are not available at the deployed environment, e.g., when drones can not fly due to adverse weather conditions}.
As a result, it is more challenging than using drone modality directly in the input, because the models need to cope with the different evolution patterns of the two data sources.

In reality, if drone data are not available either in training or testing, and the \modified{accurate segment speeds can only be approximately estimated using} trajectories of taxis or ride-hailing vehicles.
Although it is still possible to obtain segment speed and regional speed labels, the data quality will be much lower compared to drones, due to low penetration rates.
This scenario is denoted by HiMSNet\_PN(ld-), as the drone modality is considered to be absent in both training and testing phases.
For implementation, the noise in training labels has zero mean and doubled standard deviation at 0.3. 

\subsection{Experiment Results with Partial and Noisy Data}
\label{subsec: results with partial and noisy data}

\begin{table*}[!t]
    \centering
    \caption{Prediction errors with partial observation (P) and noisy (N) data}
    \resizebox{\textwidth}{!}{%
    \begin{tabular}{lcrccrc|crccrc}
        \toprule
        \multirow[c]{2}{*}{Settings}& \multicolumn{3}{c}{Segment 15 min} & \multicolumn{3}{c|}{Segment 30 min} & \multicolumn{3}{c}{Regional 15 min} & \multicolumn{3}{c}{Regional 30 min} \\
        \cmidrule(lr){2-4} \cmidrule(lr){5-7} \cmidrule(lr){8-10} \cmidrule(lr){11-13}
                      & MAE  & MAPE*   & RMSE & MAE  & MAPE*   & RMSE & MAE  & MAPE*   & RMSE & MAE  & MAPE*   & RMSE \\
        \midrule
        HiMSNet\_PN            & \textbf{1.06} & \textbf{24.5\%} & \textbf{1.75} & \textbf{1.21} & 27.4\% & \textbf{1.99} & \textbf{0.32} & \textbf{10.0\%} & \textbf{0.43} & 0.44 & 15.1\% & \textbf{0.63} \\ 
        \modified{HiMSNet\_PN(TF)} & 1.07 & 24.6\% & 1.78 & 1.22 & \textbf{27.3\%} & 2.01 & \textbf{0.32} & 10.3\% & 0.45 & \textbf{0.43} & \textbf{14.6\%} & \textbf{0.63} \\
        HiMSNet\_PN/gnn        & 1.14 & 26.2\% & 1.90 & 1.32 & 29.3\% & 2.18 & 0.34 & 10.6\% & 0.47 & 0.47 & 16.1\% & 0.67 \\ 
        HiMSNet\_PN(drone)     & 1.07 & 24.7\% & 1.77 & 1.22 & 27.6\% & 2.01 & \textbf{0.32} & 10.1\% & 0.44 & 0.44 & 15.4\% & 0.64 \\ 
        HiMSNet\_PN(ld)        & 1.12 & 26.0\% & 1.83 & 1.28 & 28.8\% & 2.07 & 0.41 & 13.3\% & 0.57 & 0.52 & 18.6\% & 0.76 \\
        HiMSNet\_PN(ld-)       & 1.16 & 25.8\% & 1.83 & 1.32 & 28.6\% & 2.07 & 0.56 & 18.6\% & 0.86 & 0.72 & 25.0\% & 1.07 \\
        \bottomrule
    \end{tabular}%
    }
    \label{tab: performance with noise and partial observation}%
\end{table*}%

Table~\ref{tab: performance with noise and partial observation} presents the prediction errors of HiMSNet under the partial observation and noisy data settings, best results in \textbf{bold}.
When compared with their \textit{full-information} counterparts in Table~\ref{tab: comparison with baseline methods}, the PN group generally has higher prediction errors, which is expected as the model has less information and more noises to deal with.

\modified{The Transformer-based~\cite{vaswani2017attention} variant HiMSNet\_PN(TF) still shows very similar performance as the default HiMSNet with GCNConv~\cite{kipf2016semi}, which again justifies the choice over a simpler design.}
Among all the other PN settings, the model with both data modalities HiMSNet\_PN has the best performance, followed by the drone-only model HiMSNet\_PN(drone) and loop detector-only model HiMSNet\_PN(ld).
This result suggests that the drone data is a better source for urban traffic forecasting, and the loop detector data can still provide complementary information when the drone data is only partially available.
However, under the \textit{full-information} setting, the loop detector data shows a slightly negative effect as the single-modality setting with drone data outperforms the two-modality default setting on the segment-level task.
Besides, the model with partial and noisy loop detector data HiMSNet\_PN(ld) does better than its counterpart with full loop detector data HiMSNet(ld).
\modified{Although loop detector data is less effective than drone data when presented alone, its complementary coverage can still provide useful information when combined with the drone data under partial observation.}
The importance of spatial correlation manifests when the data is partial and noisy.
In the segment-level task, the accuracy difference between HiMSNet\_PN and HiMSNet\_PN/gnn is much larger than the difference between HiMSNet and HiMSNet/gnn, which highlights the necessity of the GME module in the model.

\modified{While HiMSNet\_PN(ld) has training labels base-on drone data, HiMSNet\_PN(ld-) has no such high-quality training data.
Thus HiMSNet\_PN(ld-) has higher prediction errors, especially in the regional task.}
\modified{Nevertheless, the fact that HiMSNet\_PN(ld) out performs the LA method in Table~\ref{tab: comparison with baseline methods} also justifies the effects of loop detector modality. 
Because HiMSNet\_PN(ld) is trained to predict future segment speed, and by using past point speed as hints at test time, the model can infer the segment speed better than taking an average over historical data.}

\begin{figure*}[!t]
    \centering
    \includegraphics[width=\textwidth]{"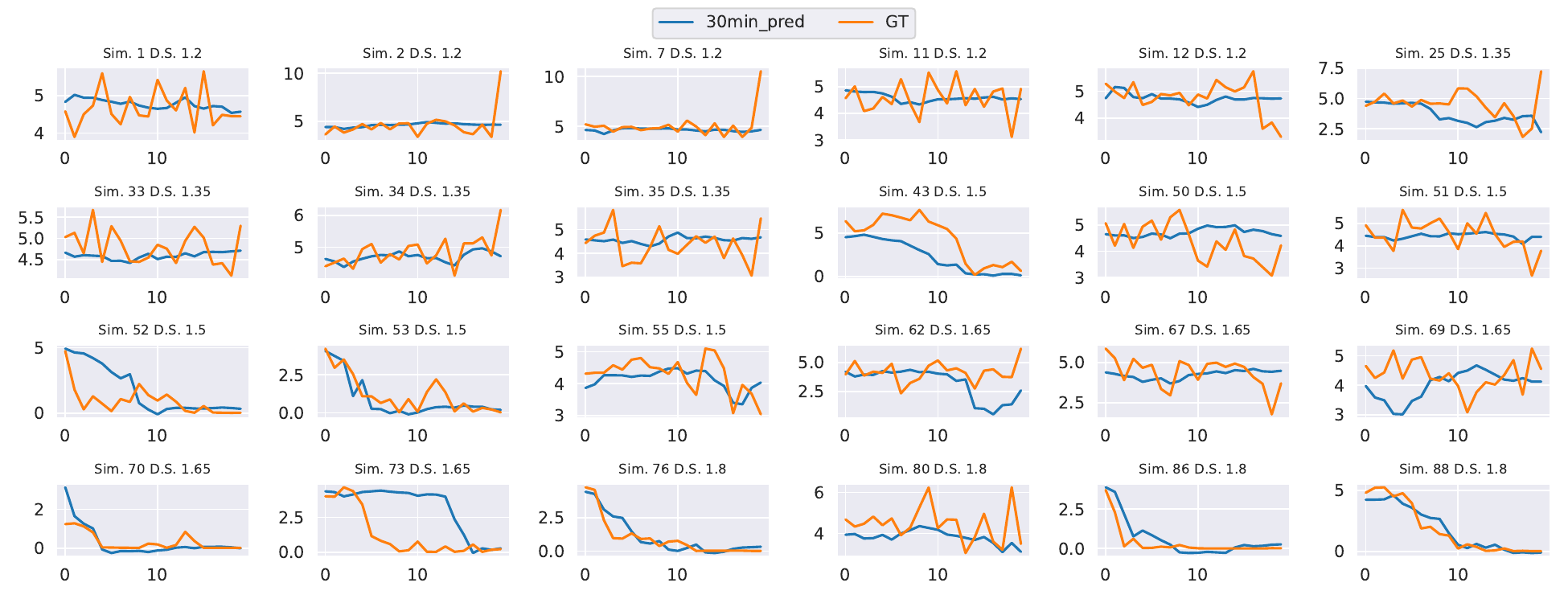"}
    \caption{30-min-ahead prediction for the road segment with ID 9971 under 10\% sensor coverage case.}
    \label{fig: 30-min-ahead prediction 9971 cvg10}
\end{figure*}

Figure~\ref{fig: 30-min-ahead prediction 9971 cvg10} shows the 30-min-ahead prediction of HiMSNet\_PN for the road segment with ID 9971 (same as Figure~\ref{fig: 30-min-ahead prediction 9971}) under the 10\% sensor coverage.
Compared to the model with full sensor coverage, the model can still predict the overall trends of the traffic states, but its predictions generally deviate more from the ground truth, as the task has become more challenging.
For example in session 52, the model predicts the speed drop later than the full-coverage case, and in session 69 the model has predicted a false congestion around time step 3. 

\begin{figure}[!t]
    \centering
    \includegraphics[width=0.4\textwidth]{"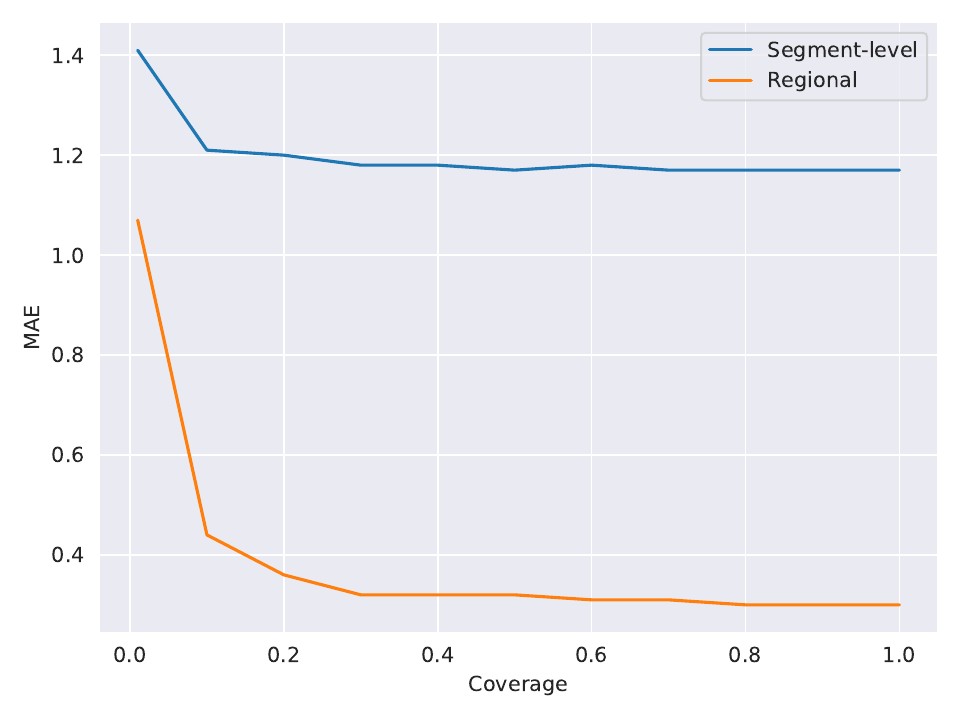"}
    \caption{30-min-ahead prediction MAE under different sensor coverage}
    \label{fig: mae by coverage}
\end{figure}

Sensor coverage is a critical factor for prediction performance.
Figure~\ref{fig: mae by coverage} shows the 30-min-ahead prediction MAE when the sensor coverage for both modalities varies simultaneously from 1\% up to 100\%.
Firstly, it is straightforward that the higher the sensor coverage, the lower the prediction errors.
Still, because of the ability to make use of correlations in space, time and modalities, HiMSNet is able to achieve similar performance for both the regional and segment-level tasks with only 20\% sensor coverage, when comapred to the full coverage case.
The error metrics of 10\% sensor coverage make it a fair trade-off choice, if the coverage is further reduced down to 1\% coverage, the performance of the model is close to the Label Average baseline in Table~\ref{tab: comparison with baseline methods}.
This is consistent with the intuition that when the model nearly has no information in the testing phase, the best it can do is to estimate a value based on the training samples, and such predictions (although not always constant) turn out to have nearly constant predictions and similar errors as the constant value used by Label Average.

\section{Conclusion}
\label{sec: conclusions}

This paper investigates the problem of multi-source traffic forecasting, i.e., utilizing drone and loop detector data for road segment-level and regional traffic speed prediction.
A flexible baseline model HiMSNet is designed to learn the dependencies in temporal, spatial and data modality dimensions.
A well-calibrated simulation dataset SimBarca is created for this study, and the experiments demonstrate the effectiveness of drone data in urban traffic forecasting.
The high-quality aerial data supports more accurate traffic forecasting compared to using loop detector data.
High traffic demand is found to be a challenging factor that causes more dynamic traffic patterns, and thus increases the difficulty of traffic forecasting.
Our experiments have shown the importance of spatial correlation and have highlighted the necessity of the drone data modality in traffic forecasting from various perspectives, especially in a real-world scenario with partial and noisy data.

In the future, this work can be improved in several directions.
First, more advanced neural network architectures can be explored to better capture the complex dependencies in the traffic data, as the current model is designed to be concise and serve as a baseline.
Second, the drone flight in the partial and noisy case can be carefully designed or intelligently optimized to provide best possible observation for traffic forecasting.
Third, more data modalities could be included, such as vehicle flow and density. 
These data are also practically available and can be integrated into the model.
\modified{
Besides, real-world experiments can be conducted to integrate the prediction approach with real-time traffic data platforms and develop deliverable services for traffic management.
Furthermore, hierarchical clustering can reduce the computational cost of the model and facilitate the deployment of the model in real-world applications.}
Finally, we hope our findings can pave the way for more accurate and reliable traffic forecasting in complex urban transportation systems.

\bibliography{references}
\bibliographystyle{IEEEtran}

\newpage

\begin{IEEEbiography}[
    {\includegraphics[width=1in,height=1.25in,clip,keepaspectratio]{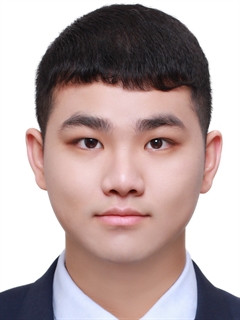}}
]{Weijiang Xiong} 
completed his bachelor's degree at Tongji University, China, and obtained his master's degree from Aalto University, Finland.
He is currently a Ph.D. student at the Urban Transport Systems Laboratory (LUTS) of EPFL, Switzerland.
His research interests are machine learning, intelligent transportation systems and smart urban mobility.
\end{IEEEbiography}

\begin{IEEEbiography}[
    {\includegraphics[width=1in,height=1.25in,clip,keepaspectratio]{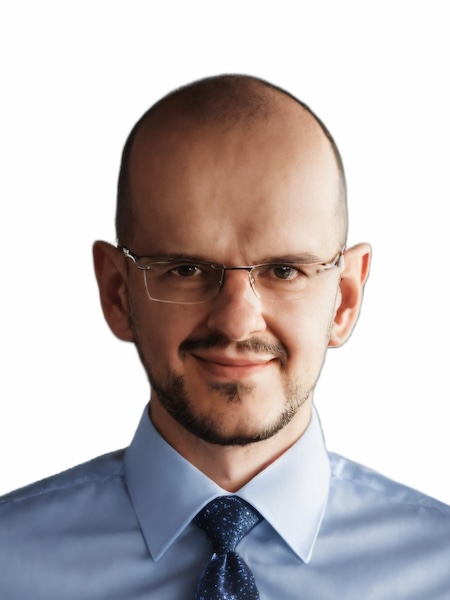}}
]{Dr. Robert Fonod} (S'10-M'15-SM'23) received his B.Sc. and M.Sc. in Cybernetics from the Technical University of Košice, Slovakia, in 2009 and 2011, respectively, and his Ph.D. in Automatic Control from the University of Bordeaux, France, in 2014. He also earned a Master's degree in Computer Science from the University of Illinois Urbana-Champaign (UIUC) in 2021. He is currently a Research Associate at EPFL, and has previously worked as a Research Scientist at the French-German Research Institute of Saint-Louis (ISL), an Assistant Professor at TU Delft, and a Postdoctoral Researcher at the Technion. He held visiting appointments at ESA and Thales Alenia Space. His research interests include guidance, navigation, and control (GNC) of aerospace vehicles, fault diagnosis, estimation and tracking, and the application of computer vision and machine learning to ITS. He currently serves as an Associate Editor for the IEEE Transactions on Aerospace and Electronic Systems.
\end{IEEEbiography}

\begin{IEEEbiography}[
    {\includegraphics[width=1in,height=1.25in,clip,keepaspectratio]{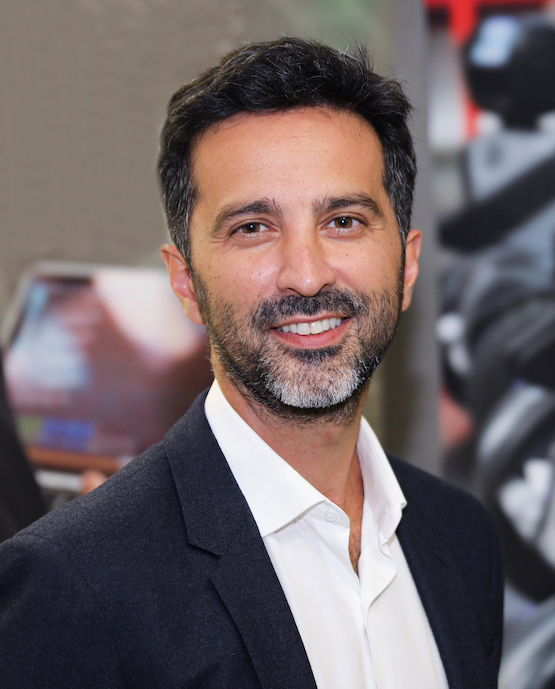}}
]{Prof. Alexandre Alahi} 
(Member, IEEE) is currently an Associate Professor at EPFL leading the Visual Intelligence for Transportation Laboratory (VITA). 
Before joining EPFL in 2017, he spent multiple years at Stanford University as a Post-doc and Research Scientist.
His research lies at the intersection of Computer Vision, Machine Learning, and Robotics applied to transportation and mobility.
\end{IEEEbiography}

\begin{IEEEbiography}[
    {\includegraphics[width=1in,height=1.25in,clip,keepaspectratio]{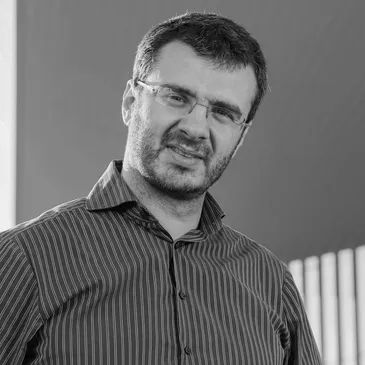}}
]{Prof. Nikolas Geroliminis} 
received the Diploma degree in civil engineering from NTUA, Greece, and the
M.Sc. and Ph.D. degrees in civil engineering from the University of California at Berkeley. 
He is currently a Full Professor with the École Polytechnique Fédérale de Lausanne (EPFL) and the Head of the Urban Transport Systems Laboratory (LUTS). 
His research interests focus primarily on urban transportation systems, traffic flow theory and control, public transportation and on-demand transport, car sharing, optimization, MFDs, and large scale networks. 
He currently serves as the Editor-in-Chief of Transportation Research Part C: Emerging Technologies journal.
\end{IEEEbiography}

\vfill

\end{document}